\documentclass{article}

 \usepackage[main, final]{neurips_2026}
\usepackage[utf8]{inputenc} % allow utf-8 input
\usepackage[T1]{fontenc}    % use 8-bit T1 fonts
\usepackage{hyperref}       % hyperlinks
\usepackage{graphicx}
\usepackage{url}            % simple URL typesetting
\usepackage{booktabs}       % professional-quality tables
\usepackage{amsfonts}       % blackboard math symbols
\usepackage{nicefrac}       % compact symbols for 1/2, etc.
\usepackage{microtype}      % microtypography
\usepackage{xcolor}         % colors
\usepackage{bm}
\usepackage{enumitem}
\usepackage{amsmath}
\usepackage{amssymb}
\usepackage{mathtools}
\usepackage{amsthm}
\usepackage{wrapfig}

\usepackage{algorithm, algpseudocode}
\usepackage{makecell}
\usepackage{multirow}
\usepackage{colortbl}
\usepackage{caption}
\usepackage{tcolorbox}
\newtheorem{theorem}{Theorem}
\newcommand{\gc}{\cellcolor[gray]{0.9}}
\title{Seeing Through the Chain: Understanding and Mitigating Hallucinations in Multimodal \\ Large Reasoning Models}

\author{
 \textbf{Hao Fang\thanks{Equal Contribution}\ \ \textsuperscript{1}},
 \textbf{Jinyu Li$^*$\textsuperscript{1,2}},
 \textbf{Jiawei Kong\textsuperscript{1}},
 \textbf{Tianqu Zhuang$^*$\textsuperscript{1}},
 \textbf{Kuofeng Gao\textsuperscript{1}}, 
 \\
 \textbf{Bin Chen\thanks{Corresponding Author}\ \ \textsuperscript{2}},
 \textbf{Shu-Tao Xia \textsuperscript{1}}
\\
 \textsuperscript{1}Tsinghua Shenzhen International Graduate School, Tsinghua University,\\
 \textsuperscript{2}Harbin Institute of Technology, Shenzhen,
\\
 \small{
   \texttt{\{fangh25, kjw25, zhuangtq23, gkf21\}@mails.tsinghua.edu.cn, }} \\
   \small{\texttt{2023311402@stu.hit.edu.cn, chenbin2021@hit.edu.cn}} \\
      \small{\texttt{xiast@sz.tsinghua.edu.cn}}
 }

\begin{document}

\maketitle

\begin{abstract}

While multimodal large reasoning models (MLRMs) have exhibited impressive capabilities, they remain prone to hallucinations, and effective solutions remain underexplored.
In this paper, we investigate the underlying causes of hallucinations in MLRMs and accordingly propose a reasoning-centric training framework for mitigation.
Specifically, we find that introducing reasoning mechanisms exacerbates models' reliance on language priors and overlooks visual inputs, leading to CoTs with reduced visual cues but redundant text tokens. Guided by the Information Bottleneck (IB) principle, we propose selectively filtering redundant thinking tokens to obtain a more compact, signal-efficient CoT that preserves task-relevant information while suppressing noise. 
We also observe that the quality of the reasoning trace largely determines whether hallucination emerges in final answers. To leverage this, we introduce a reasoning-enhanced preference optimization scheme that constructs training pairs using high-quality AI feedback.
We further propose generating high-quality negative samples for contrastive optimization via a multimodal hallucination-inducing mechanism that exposes model hallucination behaviors through carefully crafted visual and textual inducers. By modeling CoT chains as intermediate representations between inputs and final answers, we establish an IB-based theoretical justification for our method. Extensive experiments reveal consistent hallucination reduction across diverse MLRMs and benchmarks.

% \keywords{Multimodal Reasoning Models \and Hallucination Mitigation \and Reasoning-Oriented Enhancement}
% Inspired by Information Bottleneck theory,
% analyze the underlying hallucination cause and propose

% Our theoretical analysis justifies the superiority of the proposed method. Extensive experiments validate the effectiveness of C3PO across diverse MLRMs and benchmarks.

% reasoning-enhanced preference tuning with explicit positive and negative supervision. Specifically, corrected intermediate reasoning generated by a teacher model is used to construct positive training samples, while multimodal inputs that reliably induce hallucinated responses are constructed as negative samples for contrastive learning.
% while reducing attention to visual input. 
% release their unnecessary attention and .
% steer the model’s attention back to visual input
% and encouraging more visually grounded responses. 
% This also brings computational benefits by evicting redundant tokens.
% To bridge this gap, we propose C3PO, the first training-based hallucination mitigation framework for MLRMs, which incorporates our designed Chain-of-Thought Compression and Contrastive Preference Optimization techniques.
% allocate even less attention to visual tokens than their non-reasoning base models, causing them 
% Notably, pruning redundant CoT tokens also yields computational benefits, improving the average inference speed of R1-OneVision by 1.xxx across diverse tasks.
\end{abstract}
% 图怎么画？画一个幻觉的示意图，还是画MLRM和其base model的幻觉对比？
% 1. 近期研究在LRM使用meticulous reasoning，。受益于xxx上微调或者多模态强化学习，多模态推理模型[]在很多任务上取得了非常好的性能，比如xxx。
% 2. 如图1(a)所示，MLRM面临着严峻的幻觉问题，。一些MLRM甚至比其对应的基座模型还差。这很反直觉，按理来说有了推理过程，应该能够给出更加缜密且细致的分析。现有针对MLRM的幻觉算法研究较少，more thinking旨在评估现有MLRM的幻觉并调查相关的影响因子，MIRAGE隔离出了一类幻觉（reasoning hallucination）并进行分析，就是在推理时的自我逻辑错误，它重点在评估和缓解这一类幻觉。没有针对MLRM机制专门分析来改善模型在通用多模态任务上的的工作。
% 3. 我们基于推理模型的特质做了两个相关实验，观察到了两个现象，提出了两个解决针对思维链本身技术作为方案。
% 4. 在各个模型和多个幻觉Bench上取得了很好的效果，证明了所提出技术的有效性。同时由于压缩了CoT token，还带来了性能上的优势，比如能在xxx提升了百分之xx的推理速度。
\section{Introduction}
%%%%%%%%%%%%%%%%%%%%%%%%%%%%%%%%%

%%%%%%%%%%%%%%%%%%%%%%%%%%%%%
% Benefiting from multi-step reasoning mechanisms, largely benefiting from training on massive chain-of-thought (CoT) corpora containing multi-step reasoning
% practical and complex 
Large Reasoning Models (LRMs) such as DeepSeek-R1 \cite{guo2025deepseek} have recently demonstrated remarkable problem-solving capabilities across a wide range of scenarios. Motivated by these advances, numerous studies have sought to extend such powerful multi-step reasoning mechanisms to multimodal models by applying supervised fine-tuning (SFT) or reinforcement learning (RL) algorithms. Benefiting from meticulous chain-of-thought (CoT) supervision, Multimodal Large Reasoning Models (MLRMs) have achieved impressive performance on various challenging tasks such as visual mathematics \cite{wang2025sota}.

\begin{figure*}[t]
\begin{center}
\includegraphics[width=\linewidth]{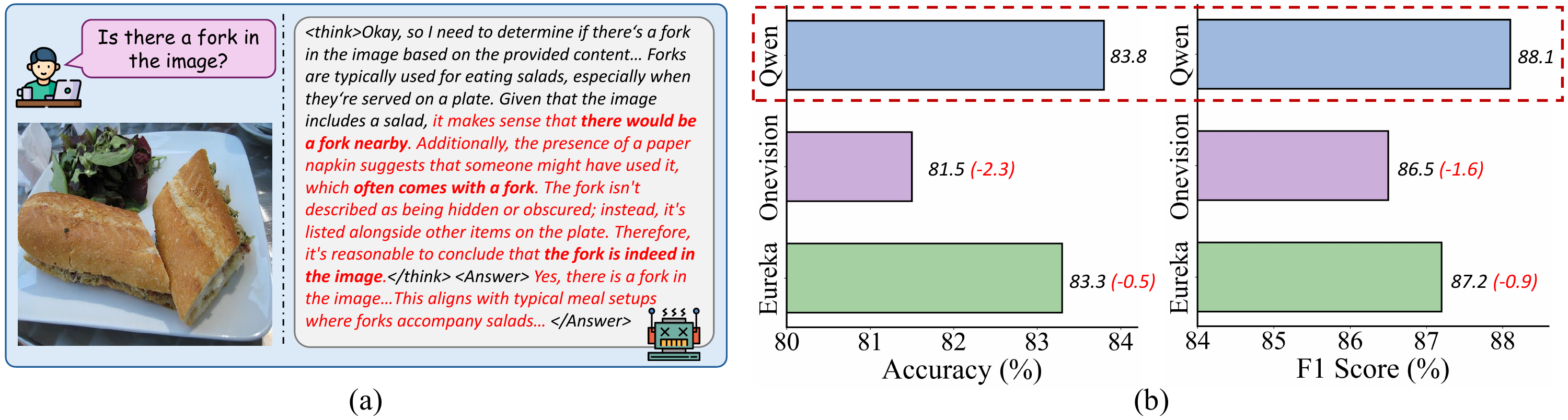}
\end{center}
\vspace{-0.7em}
\caption{(a) A hallucination case from R1-Onevision \cite{yang2025r1}. (b) Performance of MLRMs R1-Onevision and MM-Eureka \cite{meng2025mm}, and the base model Qwen2.5-VL-7B \cite{bai2025qwen2} on AMBER \cite{wang2023llm}, where \textit{high values indicate fewer hallucinations}. The hallucination increases from the base model to reasoning variants.}
 % The dashed box denotes the base model. 
\label{fig:intro}
\vspace{-0.5em}
\end{figure*}

Despite recent progress, the hallucination remains a persistent issue for MLRMs \cite{liu2025more}, \textit{i.e.}, the model occasionally generates semantically plausible yet factually incorrect or visually irrelevant content, as in Figure \ref{fig:intro}(a). 
More strikingly, evaluation of representative MLRMs and the base model reveals that MLRMs can hallucinate even more severely than their non-reasoning counterparts (Figure \ref{fig:intro}(b)). This is a counterintuitive phenomenon, as the presence of logical, structured reasoning should, in principle, facilitate a more detailed perception of the image and hence yield more faithful and image-grounded answers. These observations underscore the need for a deeper understanding of hallucinations specific to MLRMs and the development of effective mitigation strategies.
However, existing studies primarily focus on constructing benchmarks to better evaluate MLRM's hallucinations and influential factors \cite{liu2025more}, or exclusively discuss the self-contradictory reasoning within CoT traces \cite{dong2025mirage}, leaving approaches for hallucination mitigation in general multimodal tasks largely underexplored.

In this work, we first conduct two analysis tailored to the unique characteristics of MLRMs and observe two key factors relevant to their hallucinations. Based on these insights, we propose C3PO, a two-stage hallucination mitigation framework specifically designed for MLRMs via \textbf{C}oT \textbf{C}ompression and \textbf{C}ontrastive \textbf{P}reference \textbf{O}ptimization.
We begin by analyzing the attention distribution of MLRMs and observe that introducing explicit reasoning mechanisms causes models to allocate even less attention to visual tokens than their non-reasoning counterparts. 
% As a result, MLRMs tend to over-rely on language priors rather than grounding in visual inputs. 
This issue weakens visual signals in the generated CoTs while introducing redundant low-information text tokens. From the Information Bottleneck (IB) perspective \cite{IB提出,IB深度}, we model the reasoning chain as an intermediate representation between visual inputs and final answers, which should preserve critical visual cues relevant to the answer while discarding redundant or noisy content. 
Therefore, we propose to selectively filters low-importance reasoning tokens to produce a more compact and signal-efficient representation, which is theoretically justified under the IB framework.  
% and empirically improves visual grounding and answer accuracy.
% To understand this phenomenon, we revisit the reasoning chain from the perspective of Information Bottleneck (IB) \cite{IB提出,IB深度}.  The CoT chain serves as an intermediate representation between visual input and the final answer. To encourage the answer is more visually grounded, the CoT chain hence should retain task-relevant visual information while discarding noisy and redundant details. 
% Motivated by recent findings \cite{xia-etal-2025-tokenskip, li2025thinkless} that the reasoning chain is highly redundant, we propose a simple yet effective approach that filters out a portion of low-importance reasoning tokens to release their attention and redirect the model’s focus back to the visual input. This is also theoretically supported by the Information Bottleneck (IB) principle \cite{IB提出,IB深度}, as removing low-information reasoning tokens suppresses noise within the intermediate representation between queries and final responses, hence strengthening the information flow. 
% We achieve this by fine-tuning the MLRM directly on data samples consisting of pre-pruned CoTs and the model’s original, unmodified answers. 
Remarkably, this strategy directly reduce hallucinations without introducing any additional prior information.
% To address this issue, we propose a simple strategy that removes some redundant reasoning tokens based on their importance scores. This is because not all tokens in the reasoning chain contribute equally to the prediction \cite{}, with some being redundant and consuming unnecessary attention. This removement operation helps redirect attention back to the crucial visual tokens. Interestingly, fine-tuning directly on data pairs consisting of the pruned CoTs and the model’s original, unmodified answers already leads to a significant reduction in hallucinations, demonstrating the practical effectiveness of this approach.

Next, we investigate the causal relationship of hallucinations from reasoning chains to final answers.
Our analysis reveals a key finding that high-quality CoTs consistently lead to hallucination-free responses, whereas hallucinated CoTs substantially amplify the likelihood of hallucinations in final answers. 
Building on this, we propose a reasoning-enhanced preference learning strategy based on direct preference optimization (DPO). Specifically, we leverage expert feedback from advanced MLLMs to enhance the quality of model-generated CoTs. Data samples with enhanced traces are used as preferred samples, while the original outputs serve as rejected ones. To expose the model’s intrinsic hallucination behaviors, we further introduce a multimodal hallucination-inducing mechanism that carefully crafts visual and textual inputs to elicit hallucination patterns as additional negative contrasts. By performing contrastive preference tuning on these crafted pairs, the model is trained to generate more reliable and coherent reasoning chains, substantially reducing hallucinations in final answers. 

% augmented with a reasoning-centric objective
% In summary, our contributions are as follows:
%  \begin{itemize}
%     \item We first experimentally uncover the hallucination mechanisms specific to the unique reasoning process of MLRMs and identify two key challenges that guide our reasoning-oriented mitigation strategy.
%     \item We propose C3PO, a two-stage hallucination mitigation framework for MLRMs that first leverages SFT to compress redundant CoT tokens, and then performs a reasoning-enhanced contrastive preference optimization (CPO) using high-quality preference pairs derived from AI feedback.
%     \item We further introduce a multimodal hallucination-inducing technique that exploits the inherent failure behaviors of MLRMs via well-designed inducers to obtain informative contrastive references.
%     \item We develop a theoretical framework from an information-bottleneck perspective that provides principled justification for the superiority of C3PO. Extensive experiments on various MLRMs across a wide range of benchmarks demonstrate that the proposed framework substantially reduces MLRMs' hallucinations
% \end{itemize}
\textbf{Contributions.} We first uncover the hallucination causes specific to the reasoning mechanism in MLRMs and identify two key challenges that motivate the design of C3PO, a two-stage hallucination mitigation framework for MLRMs. C3PO first leverages SFT to obtain more information-dense CoTs, and then performs the reasoning-enhanced contrastive preference optimization (CPO). In addition, we introduce a multimodal hallucination-inducing technique that exploits their inherent failure behaviors via well-designed inducers to generate contrastive references. Furthermore, we develop a theoretical framework from an IB perspective that provides principled justification for C3PO's superiority.

To validate the effectiveness, we conduct extensive experiments on various MLRMs across multiple benchmarks, which demonstrate that the proposed framework substantially reduces hallucinations. 

\textbf{Related work.} Due to page limits, we discuss related studies in Appendix \ref{sec:related_work}.
\section{Understanding Hallucination in MLRMs}

This section provides preliminary analyses tailored to the unique characteristics of MLRMs to inspire the principled design of our framework.
A key distinction between MLRMs and conventional multimodal large language models (MLLMs) is the explicit incorporation of CoT reasoning, based on which we investigate two fundamental questions closely related to hallucination behaviors: 
\begin{itemize}[leftmargin=*, noitemsep, topsep=0pt]
    \item \textbf{Q1:} \textit{How does the introduction of an additional reasoning process influence models' attention allocation? Does reasoning encourage more attention to visual inputs or instead shift reliance toward language priors due to the increased number of textual tokens?} 
    % ?}
    \item \textbf{Q2:} \textit{Does hallucination arising in the reasoning chain causally propagate to the final answer due to the autoregressive nature of LLMs? More importantly, does a hallucination-free reasoning chain necessarily guarantee a hallucination-free final answer?}
\end{itemize}

\begin{figure*}[t]
\begin{center}
\includegraphics[width=\linewidth]{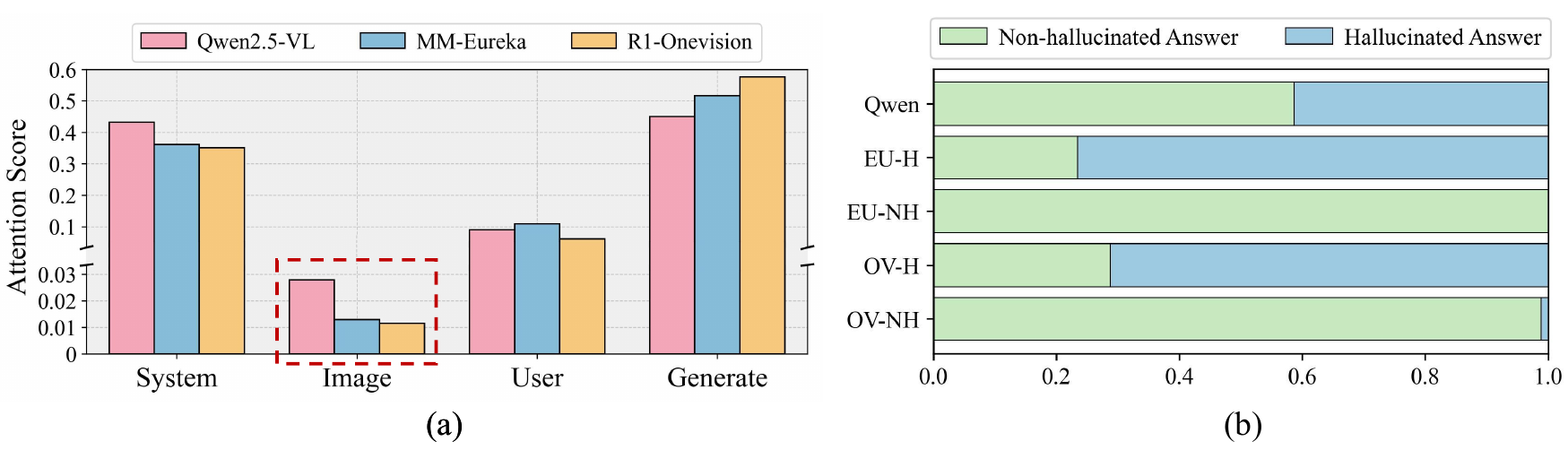}
\end{center}
\vspace{-0.5em}
\caption{Two important hallucination analyses for MLRMs. (a) Attention distributions of MLRMs and the non-reasoning Qwen2.5-VL on 300 samples from MSCOCO \cite{lin2014microsoft}. As observed, MLRMs exhibit a substantial reduction in attention to visual tokens. (b) Proportion of hallucinated answers given hallucinated (-H) and non-hallucinated (-NH) CoTs under CHAIR \cite{rohrbach2018object} evaluation.
EU denotes MM-Eureka, and OV denotes R1-Onevision. The result of Qwen2.5-VL is reported as a reference.}
\label{fig:attention_cot}
\vspace{-0.5em}
% base model 
% while a prominent increase to textual token
% TODO；如果不显著，那就选最长的那些来展示，在caption里面说明我们选了最显著的50条平均来说明此现象。
\end{figure*}

\textbf{Question 1}. Previous studies \cite{favero2024multi, fang2025grounding} have shown that a primary cause of hallucinations is models' tendency to allocate excessive attention to text tokens while ignoring critical visual inputs. As a result, the responses are more influenced by the language priors rather than the actual image input.
% verbose reasoning chains 

The attention analysis in Figure~\ref{fig:attention_cot}(a) shows that reasoning models further exacerbate attention bias. Compared to their non-reasoning base models, MLRMs significantly decrease attention to visual tokens while allocating even more attention to textual tokens. This shift amplifies the imbalance between visual grounding and language priors during generation, increasing the risk of hallucinations.

\textbf{Summary 1.} Introducing an additional reasoning mechanism decreases focus on visual tokens and amplifies reliance on textual priors, which can produce CoTs with diluted visual information and redundant low-information text tokens. 
Building on the IB principle and empirical findings that many CoT tokens are redundant and irrespective of model decisions \cite{xia-etal-2025-tokenskip, feng2025efficient}, we propose a selective CoT token pruning strategy to enhance information flow and encourage reliable answers.

% This observation motivates the removal of redundant reasoning tokens to restore attention to visual inputs. 

\textbf{Question 2.} 
The CoT chain plays a critical role in shaping the final answer. 
% Intuitively, due to the auto-regressive nature of language models, 
To investigate the potential causal relationship between hallucinations in reasoning chains and final answers, we decouple these two sources of hallucination under the CHAIR benchmark and conduct a causal analysis.

Figure~\ref{fig:attention_cot}(b) shows that when the CoT is hallucinated, most generated answers also contain hallucinated content, with a higher hallucination rate compared with the non-reasoning model. However, a non-negligible portion of hallucination-free answers remains, suggesting that hallucinations in CoTs mostly, but not always, propagate to the final answers.
% This confirms that hallucinations in the CoT chain can strongly propagate to the final responses. 
% Notably, the resultant final-answer hallucination rate is significantly higher than that of the non-reasoning base model.
In contrast, we find that non-hallucinated reasoning chains almost always yield hallucination-free answers. This provides preliminary evidence that hallucinations can be effectively mitigated by improving the quality of reasoning chains.
% as a high-quality reasoning process enables more faithful analysis of the image and hence facilitates more reliable answers. 
% Another important observation is that hallucination-free reasoning chains almost always yield hallucination-free answers. This indicates that, in this regime, the final answer generation largely adheres to the underlying reasoning chain without introducing additional hallucinations. Consequently, a high-quality reasoning process enables more faithful image-grounded reasoning and leads to more reliable outputs. This provides initial evidence that improving reasoning chain quality is a promising direction for mitigating hallucinations in MLRMs.

\textbf{Summary 2.} Hallucinated CoTs increase the likelihood of hallucinations in final answers, whereas reliable reasoning chains strongly suppress such errors. Accordingly, we design a novel preference optimization approach with a reasoning-centric objective for enhanced CoT chains. 
% where we introduce AI-enhanced reasoning as positive supervision and deliberately induced hallucinated content as contrastive negatives.

% indicate that the quality of the reasoning chain plays a critical role in determining answer faithfulness,
\section{Method}
% In this section, we introduce the proposed two-stage hallucination mitigation framework for MLRMs. Moreover, we provide a theoretical analysis from an IB perspective to justify its superiority.
% which first prunes reasoning traces via SFT to facilitate information flow and then enhances reasoning quality through preference optimization with contrastive supervision.

\subsection{Chain-of-Thought Compression}
Based on our preliminary analysis, we aim to remove low-information reasoning tokens from the reasoning chain to obtain a more compact and informative intermediate representation, which aligns with the IB principle \cite{IB提出} and is also motivated by the over-thinking issue in reasoning models \cite{chen2024not, li2025thinkless}.
% where excessively long CoTs not only increase computational burdens but also degrade model performance. 
Inspired by recent findings that LLMs can perform reliably with pruned CoTs \cite{xia-etal-2025-tokenskip}, we implement compression by training the model to reason and predict based on trimmed CoT traces.
% This is also consistent with the IB principle \cite{IB提出} and further motivated by the over-thinking issue in reasoning models \cite{}, where excessively long CoTs not only increase computational burdens but also degrade performance.

\textbf{Data Construction.} Given a multimodal reasoning model $f_{\theta}$, we query $f_{\theta}$ with image–question pairs $(v, x)$ to generate reasoning chain $z$ and final answer $y$. We then identify reasoning tokens with limited impact on model predictions based on LLMlingua-2 \cite{pan2024llmlingua}, a token-level importance scoring model trained with GPT-4 annotations that has been widely adopted in compression studies \cite{pan2024llmlingua, xia-etal-2025-tokenskip}.  
Specifically, LLMlingua-2 is capable of assigning higher importance scores to semantically informative tokens (\textit{e.g.}, object attributes and scene descriptions in image captioning), while lower scores to function words and transitional tokens with limited semantics and minimal influence on model decisions (See examples in Appendix \ref{app:visual_pruned_cot}). Tokens are then ranked by importance, and the top $\gamma$ percentile is retained to form the pruned chain $z'$. We adopt a threshold of $\gamma=90\%$ to reduce reasoning redundancy while preserving sufficient information to maintain clean performance. The resulting dataset of $N$ samples is denoted as $\mathcal{D}_s = \{(x_{(i)}, v_{(i)}, z'_{(i)}, y_{(i)})\}_{i=1}^{N}$.
% where we adopt a threshold of $\gamma=90\%$ to effectively reduce hallucinations while preserving sufficient information to maintain clean performance.
% ^{\mathrm{trim}}

\textbf{Model Training.}
We then use the constructed dataset $\mathcal{D}_s$ to perform LoRA-based SFT on the MLRM, which trains the model to reason with compressed CoTs and generate faithful answers. Let $l_{z'}$ and $l_y$ denote the sequence lengths of $z'$ and $y$ respectively, the training objective can be expressed as:
\begin{equation}
\begin{split}
\mathcal{L}_{\mathrm{SFT}}
= -\mathbb{E}_{(v,x,z',y)\sim\mathcal{D}_s}
\sum_{t=1}^{l_{z'}}
\log p_\theta(z'_t \mid v, x, z'_{<t})+\sum_{t=1}^{l_y}
\log p_\theta(y_t \mid v, x, z', y_{<t}).
\end{split}
\end{equation}
Notably, this SFT process does not introduce any additional supervision information beyond the model’s own outputs. We simply prune the generated reasoning chains and pair them with the original, unmodified answers for training. As shown in Appendix \ref{app:add_exps}, this strategy already brings a notable reduction in hallucinations, confirming the rationality of our analysis and design.

\begin{figure*}[t]
\begin{center}
\includegraphics[width=\linewidth]{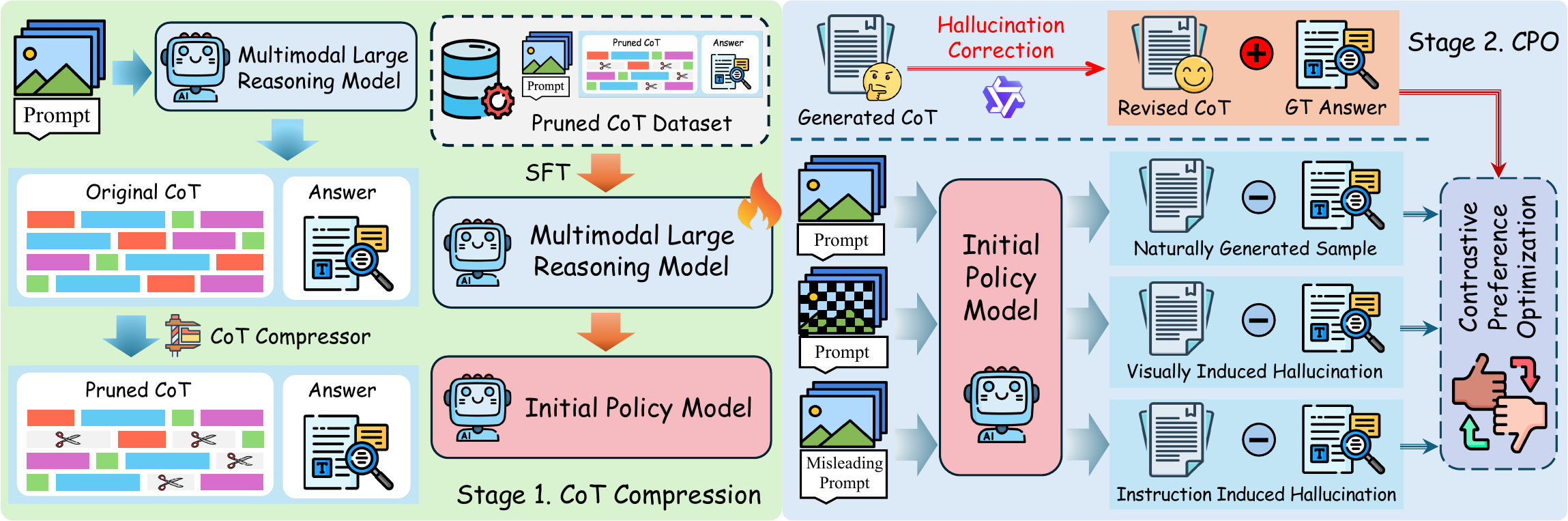}
\end{center}
\caption{Overview of the proposed C3PO framework. We first construct SFT datasets by removing redundant low-importance tokens in generated CoTs. Next, we perform hallucination-aware preference optimization, where preferred samples contain CoTs enhanced by an advanced open-source MLLM, while the original unmodified outputs serve as rejected ones. To further enhance the optimization, we propose a multimodal hallucination-inducing technique that crafts degraded visual inputs and misleading prompts to elicit MLRM's inherent hallucination patterns as negative contrasts. 
% This contrastive strategy substantially exposes the models' inherent hallucination patterns and, in turn, facilitates more faithful final responses.}
}
\label{fig:pipeline}
\end{figure*}

% Moreover, we highlight that this compression-based approach introduces additional benefits to inference speed and computational efficiency, especially for reasoning models that generally produce excessively long responses, formulating an effective and efficient method.

\subsection{Contrastive Preference Optimization}
Motivated by the preceding causal analysis, we propose a reasoning-oriented preference optimization strategy that constructs high-quality preference data pairs to effectively strengthen the model's reasoning quality, which then facilitates the generation of hallucination-free answers.
% hallucination-aware preference 
% To obtain robust CoTs as preferred samples,
% unlike existing approaches that typically rely on expensive closed-source models for annotation \cite{lee2023rlaif, yu2025rlaif}, 

To obtain reliable CoTs as preferred samples, we adopt the well-established RLAIF paradigm that constructs data by incorporating high-quality feedback from advanced MLLMs. We decompose the refinement into sentence-level hallucination detection and correction~\cite{yang2025mitigating}, which simplifies the task and enables open-source MLLMs to provide more reliable guidance.
Specifically, we condition the advanced Qwen3-VL~\cite{bai2025qwen3vltechnicalreport} on the input image $v$, the question $x$, and the ground-truth answer $y_{\mathrm{GT}}$, and instruct it to identify and correct hallucinated content in the reasoning chain $z_{\mathrm{Gen}}$ generated by the MLRM.
% The model is instructed to  in the CoT at the sentence level, to improve the faithfulness of the reasoning trace to the visual input.
Accordingly, the refined reasoning chain $z_{\mathrm{Rev}}$ is then used as the preferred sample, and the original unmodified reasoning chains $z_{\mathrm{Gen}}$ that may contain hallucinations serve as inferior ones.
% The refined reasoning chain $z_{\mathrm{Rev}}$ paired with the ground-truth answer $y_{\mathrm{GT}}$ is then used as the preferred sample. Accordingly, the original unmodified reasoning chain $z_{\mathrm{Gen}}$ and answer $y_{\mathrm{Gen}}$ that may contain hallucinations serve as inferior samples.

\textbf{Multimodal Hallucination-Inducing Contrast.} Beyond crafting high-quality positive samples, we further conduct a reverse-thinking analysis. While it is challenging to consistently derive hallucination-free responses from a given MLRM, eliciting the opposite hallucinated predictions is considerably easier. Such hallucinated samples are also highly informative as they explicitly expose the model’s intrinsic hallucination patterns, which can serve as negative references for contrastive correction.

Therefore, we propose a multimodal hallucination-inducing mechanism that carefully crafts both visual and textual inputs to elicit outputs with rich hallucinated content.
As shown in Fig.~\ref{fig:pipeline}, we induce visual hallucinations by applying random masks to input images, which diminishes substantial ground-truth visual evidence and implicitly forces the model to rely on language priors, resulting in semantically coherent yet hallucinated CoT chains and answers $(z_{\mathrm{img}}, y_{\mathrm{img}})$.
For text modality, we design a misleading prompt (see Appendix \ref{appendix:data_construction}) that directly instructs the model to prioritize its language priors while partially ignoring the visual input to create a plausible-sounding hallucinated reasoning trace and answer $(z_{\mathrm{ins}}, y_{\mathrm{ins}})$. These two mechanisms effectively simulate MLRM's inherent hallucination behaviors, encouraging the model to perceive and correct such failure modes.
 % in real-world inference, 
% providing informative negative samples that 
% for behavior correction. With these carefully constructed negative samples as contrastive references, the model is

% While the formulation of $\mathcal{L}_{\mathrm{CPO}}$ improves both the reasoning and answer quality, the resulting preference signal is jointly influenced by the reasoning trace and the final answer, which may not consistently provide direct and sufficient reinforcement for the reasoning chain.
\textbf{Reasoning-Enhanced Optimization.}
Given the preference dataset $\mathcal{D}_p$ constructed by the above operations, we strengthen the reasoning process by introducing a reasoning-centric preference loss,
which encourages the model to favor reliable reasoning trajectories over inferior ones: 
\begin{equation}
\begin{aligned}
\mathcal{L}_{\mathrm{RE}}
= 
- \mathbb{E}_{(v,x, z_{\mathrm{Rev}},\mathcal{N}_z)\sim\mathcal{D}_p} 
&\sum_{z_{l} \in \mathcal{N}_z}\log \sigma \Big[
\beta \log
\frac{p_\theta(z_\mathrm{Rev}| v, x)}
     {p_{\theta_{\mathrm{ref}}}(z_\mathrm{Rev}|v, x)}-
\beta\log
\frac{p_\theta(z_l|v, x)}
     {p_{\theta_{\mathrm{ref}}}(z_l| v, x)}
\Big],\\
&\quad\quad\quad\quad\quad\quad\quad\quad\quad\text{where}\
\mathcal{N}_z
=
\{z_{\mathrm{Gen}},
  z_{\mathrm{img}},
  z_{\mathrm{ins}}\}.
\end{aligned}
\end{equation}
% \begin{equation}
% \begin{aligned}
% \mathcal{L}_{\mathrm{RE}}
% &= 
% - \mathbb{E}_{(v,x, z_{\mathrm{Rev}},\mathcal{N}_z)\sim\mathcal{D}_p} 
% \sum_{z_{l} \in \mathcal{N}_z} 
% \log \sigma \Big[
% \beta r_{\theta}(z_{\mathrm{Rev}} \mid v, x)
% -
% \beta r_{\theta}(z_{l}| v, x)
% \Big],
% \\
% &\quad\quad\quad\quad\text{where}\
% \mathcal{N}_z
% =
% \{z_{\mathrm{Gen}},
%   z_{\mathrm{img}},
%   z_{\mathrm{ins}}\},\ r_\theta(z\mid v, x)
% \triangleq
% \log
% \frac{p_\theta(z\mid v, x)}
%      {p_{\theta_{\mathrm{ref}}}(z\mid v, x)}.
% \end{aligned}
% \end{equation}
Here, the $\theta_{\mathrm{ref}}$ denotes the parameters of the reference model. This design provides direct preference supervision on the reasoning process, which effectively stabilizes and facilitates the generation of faithful CoT chains, hence significantly reducing hallucinations in final answers.

% $\mathcal{L}_{\mathrm{CPO}}$ as an auxiliary objective.
% This auxiliary loss
However, optimizing reasoning preferences alone does not guarantee that the outputs consistently follow the desired \textit{(reasoning, answer)} format, and may even degrade the coherence and quality of the final answer. To stabilize holistic generation behavior and explicitly improve answer quality, we further adopt a standard DPO objective that models preferences over the joint output $(z, y)$:  
\begin{equation}
\begin{aligned}
&\mathcal{L}_{\mathrm{DPO}}
= 
- \mathbb{E}_{(v,x, z_{\mathrm{Rev}}, y_{\mathrm{GT}}, \mathcal{N})\sim\mathcal{D}_p}\\
&\sum_{(z_{l},y_{l}) \in \mathcal{N}}
\log \sigma \Big[
\beta \log
\frac{p_\theta(z_{\mathrm{Rev}}, y_{\mathrm{GT}} \mid v, x)}
     {p_{\theta_{\mathrm{ref}}}(z_{\mathrm{Rev}}, y_{\mathrm{GT}} \mid v, x)}
-
\beta \log
\frac{p_\theta(z_l, y_l \mid v, x)}
     {p_{\theta_{\mathrm{ref}}}(z_l, y_l \mid v, x)}
\Big],
\\
&\quad\quad\quad\quad\quad\quad\quad\quad\text{where}\
\mathcal{N}
=
\{(z_{\mathrm{Gen}},y_{\mathrm{Gen}}),
  (z_{\mathrm{img}},y_{\mathrm{img}}),
  (z_{\mathrm{ins}},y_{\mathrm{ins}})\}.
     \label{equ:regular}
\end{aligned}
\end{equation}
% \begin{equation}
% \begin{aligned}
% &\mathcal{L}_{\mathrm{DPO}}
% = 
% - \mathbb{E}_{(v,x, z_{\mathrm{Rev}}, y_{\mathrm{GT}}, \mathcal{N})\sim\mathcal{D}_p}\\
% &\sum_{(z_{l},y_{l}) \in \mathcal{N}}
% \log \sigma \Big[
% \beta \log
% \frac{p_\theta(z_{\mathrm{Rev}}, y_{\mathrm{GT}} \mid v, x)}
%      {p_{\theta_{\mathrm{ref}}}(z_{\mathrm{Rev}}, y_{\mathrm{GT}} \mid v, x)}
% -
% \beta \log
% \frac{p_\theta(z, y \mid v, x)}
%      {p_{\theta_{\mathrm{ref}}}(z, y \mid v, x)}
% \Big],
% \\
% &\quad\quad\text{where}\
% \mathcal{N}
% =
% \{(z_{\mathrm{Gen}},y_{\mathrm{Gen}}),
%   (z_{\mathrm{img}},y_{\mathrm{img}}),
%   (z_{\mathrm{ins}},y_{\mathrm{ins}})\},
% \\
% &\quad\quad\quad\quad\quad\quad\quad r_\theta(z, y \mid v, x)
% \triangleq
% \log
% \frac{p_\theta(z, y \mid v, x)}
%      {p_{\theta_{\mathrm{ref}}}(z, y \mid v, x)}.
%      \label{equ:regular}
% \end{aligned}
% \end{equation}

By additionally conducting preference optimization over the holistic output distribution, we guide the model to generate well-formed and faithful responses with both robust CoTs and final answers.
% enhance both the reasoning and answer quality of the MLRM, 
Furthermore, existing studies \cite{wang2024mdpo} have observed that the probability of preferred responses may decrease because DPO essentially focuses on relative preferences.
Similarly, we introduce the anchor regularization to constrain the positive reward above a predefined threshold $\delta$:
\begin{equation}
\begin{aligned}
\mathcal{L}_{\mathrm{Anc}}
= - &\mathbb{E}_{(v,x, z_{\mathrm{Rev}},y_{\mathrm{GT}})\sim\mathcal{D}_p} \Big[\log \sigma 
\Big(\beta r_{\theta}(z_{\mathrm{Rev}} \mid v, x)
- \delta\Big)\\
&+\lambda_{\mathrm{DPO}}\log \sigma \Big(
\beta r_{\theta}(z_{\mathrm{Rev}}, y_{\mathrm{GT}} \mid v, x)
- \delta\Big)\Big],
\end{aligned}
\label{eq:anchor_loss}
\end{equation}
where $\lambda_{\mathrm{DPO}}$ balances the contributions of two loss terms. With all the above loss terms, the overall optimization objective can be formulated as:
\begin{equation}
\mathcal{L}_{\mathrm{total}} = \mathcal{L}_{\mathrm{RE}} + \lambda_{\mathrm{DPO}}\mathcal{L}_{\mathrm{DPO}}  + \lambda_{\mathrm{Anc}}\mathcal{L}_{\mathrm{Anc}},
\label{eq:total_loss}
\end{equation}

\noindent where $\lambda_{\textrm{Anc}}$ controls the influence of anchor loss term $\mathcal{L}_{\mathrm{Anc}}$.

\subsection{Theoretical Analysis}
\label{sec:理论分析}

In addition to empirical analysis, we build a theoretical framework to validate the effectiveness of C3PO. Let $X$ denote the input, $Y$ the ground truth response, and $Z$ the reasoning trace. Note that $Y\to X\to Z$ is a Markov chain, where $Z$ serves as the intermediate representation between input and output. Based on the Information Bottleneck theory \cite{IB提出, IB深度, IB开盒}, a well-trained model should:
\begin{itemize}[leftmargin=28pt]
    \item Minimize $\mathcal{I}(X;Z)$, \textit{i.e.}, the mutual information between the model input $X$ and intermediate representation $Z$. This urges the model to compress redundant information from the input, thereby improving the information density of the reasoning chain $Z$. 
    \item Maximize $\mathcal{I}(Y;Z)$, \textit{i.e.}, the mutual information between the ground truth $Y$ and intermediate representation $Z$. This encourages the model to retain task-relevant information in the CoTs, such as critical visual cues necessary for producing the correct answer.
\end{itemize}

Overall, the Information Bottleneck objective is defined as
\begin{align}
\label{eq:IB定义}
\mathcal{L}_\text{IB}(Z)=\mathcal{I}(X;Z)-\lambda_\text{IB}\mathcal{I}(Y;Z),
\end{align}
where $\lambda_\text{IB}>0$ is a hyperparameter that controls the trade-off between compression of redundant information from the input and preservation of predictive information about the ground truth. A smaller $\mathcal{L}_\text{IB}(Z)$ corresponds to a better trade-off, yielding an intermediate representation that is both compact and highly informative for prediction. In the considered scenario, we focus on the regime $\lambda_\text{IB}>1$ to prioritize accuracy over compression, as suggested in a previous study \cite{IB变分}. 

% where $\lambda_\text{IB}>0$ is a hyperparameter that controls the trade-off between compression of redundant information from the input and preservation of predictive information about the target. A smaller $\mathcal{L}_\text{IB}(Z)$ corresponds to a more favorable trade-off,  In our setting, we focus on the regime $\lambda_\text{IB}>1$ to place greater emphasis on predictive accuracy over compression, following~\cite{IB变分}.
 % In our setting, we focus on the regime $\lambda_\text{IB}>1$ to prioritize predictive accuracy over compression, following~\cite{IB变分}.
% Formally, the effectiveness of our two techniques, \textit{i.e.}, compression and correction are validated by the following theorems:

% (\textit{i.e.}, CoT compression and enhancement) 
Formally, we justify the effectiveness of our proposed two key strategies in reducing the IB objective by the following two theorems, respectively:

\begin{theorem}[Compression reduces IB]
\label{thm:压缩降低IB}
Let the reasoning chain $Z$ be decomposed into $Z_{\mathrm{ret}}$ and $Z_{\mathrm{trim}}$, where $Z_{\mathrm{ret}}$ denotes the retaining task-critical part, $Z_{\mathrm{trim}}$ represents the trimmed redundant part, $Z_{\mathrm{ret}}\cup Z_{\mathrm{trim}}=Z$ and $Z_{\mathrm{ret}}\cap Z_{\mathrm{trim}}=\emptyset$. Then, the IB objective satisfies
\begin{equation}
\mathcal{L}_\textnormal{IB}(Z)\ge\mathcal{L}_\textnormal{IB}(Z_{\mathrm{ret}}).
\end{equation}
\end{theorem}

\begin{theorem}[Enhancement reduces IB]
\label{thm:增强降低IB}
Let the reasoning chain $Z$ be refined to $Z'$ via our reasoning-enhanced optimization, which supplements additional information about the ground truth $Y$.
Then, the IB objective satisfies
\begin{equation}
\mathcal{L}_\textnormal{IB}(Z) > \mathcal{L}_\textnormal{IB}(Z').
\end{equation}
\end{theorem}
Detailed proofs are provided in Appendix~\ref{sec:证明}.
Theorems~\ref{thm:压缩降低IB}~and~\ref{thm:增强降低IB} reveal that both compressing and enhancing the CoT reduce the model's information bottleneck objective, which facilitates a more compact and faithful intermediate reasoning, thereby leading to more reliable final answers.

% $\mathcal{D}=\left\{(x^{(i)}, v^{(i)}, z'^{(i)}, y^{(i)}) \right\}_{i=1}^{N}.$

% \begin{equation}
% \mathcal{L}_{\mathrm{SFT}}(\theta)
% =
% - \mathbb{E}_{(v,x,z,y)\sim\mathcal{D}}
% \left[
% \sum_{t=1}^{T_z}
% \log p_\theta(z_t \mid v, x, z_{<t})
% +
% \sum_{t=1}^{T_y}
% \log p_\theta(y_t \mid v, x, z, y_{<t})
% \right].
% \end{equation}

% \begin{equation}
% \begin{split}
% &\mathcal{L}_{\mathrm{PO}}
% = -\mathbb{E}_{(v,x,z_{\mathrm{Rev}},y_{\mathrm{GT}},z_{\mathrm{Gen}},y_{\mathrm{Gen}},z_{\mathrm{img}},y_{\mathrm{img}}, z_{\mathrm{ins}},y_{\mathrm{ins}})\sim\mathcal{D}} \\
% &\Bigg[
% \log \sigma \Big(
% \beta r_{\theta}(z_{\mathrm{Rev}},y_{\mathrm{GT}}|v, x)-\beta r_{\theta}(z_{\mathrm{Gen}},y_{\mathrm{Gen}}|v, x)
% \Big) 
% \\
% &+
% \log \sigma \Big(
% \beta r_{\theta}(z_{\mathrm{Rev}},y_{\mathrm{GT}}|v, x)-\beta r_{\theta}(z_{\mathrm{img}},y_{\mathrm{img}}|v, x)
% \Big)
% \\
% &+
% \log \sigma \Big(
% \beta r_{\theta}(z_{\mathrm{Rev}},y_{\mathrm{GT}}|v, x)-\beta r_{\theta}(z_{\mathrm{ins}},y_{\mathrm{ins}}|v, x)
% \Big)
% \Bigg],
% \\
% &\quad\quad\quad\quad\text{where}\ r_\theta(z, y \mid v, x)
% \triangleq
% \log \frac{p_\theta(z, y \mid v, x)}
% {p_{\theta_{\mathrm{ref}}}(z, y \mid v, x)}.
% \end{split}
% \end{equation}
\section{Experiments}
To reveal the superiority of our method, this section provides comprehensive experiments on various MLRMs across different evaluation benchmarks. Due to page limits, more experimental results and visualization comparisons are provided in Appendix \ref{app:add_exps} and \ref{app:visual}, respectively. The code is available at: \textcolor{magenta}{\url{https://github.com/ffhibnese/MLRM_Understanding_Mitigating_Hallucination}}.
\subsection{Experimental Setup}
\textbf{Models and Datasets.}
% 主要介绍我们考虑了哪些MLRM模型、两个训练数据集是啥，有多少个数据对，如何构造的。
To verify the effectiveness of our method, we conduct experiments on five representative MLRMs: Orsta-R1-7B \cite{ma2025one}, ThinkLite-7B \cite{wang2025sota}, MM-Eureka-7B \cite{meng2025mm}, MM-R1-7B \cite{leng2025mmr1}, and R1-Onevision-7B \cite{yang2025r1}. To construct training data, we randomly sample 20k data pairs from the RLAIF-V dataset \cite{yu2025rlaif} as seed data. The CoT compression ratio is $90\%$ by default. We apply a random masking ratio of $0.3$ to the visual input for hallucination-inducing responses. 
% (2) Preference Dataset ($\mathcal{D}_p$): Constructed by pairing a positive sample (refined compressed CoT appended to ground-truth answer) with three negative variants generated by the reference model: (i) a natural negative from standard generation; (ii) a visually induced negative elicited by applying a random masking ratio of 0.3 to the visual input tokens during inference; and (iii) an instruction-induced negative crafted by injecting a specific system prompt that explicitly steers the model towards fabrication. 
More details, such as the designed \textit{misleading prompt}, are in Appendix~\ref{appendix:data_construction}.

\textbf{Evaluation Benchmarks.}
% 介绍考虑了哪些幻觉Bench、通用性能的Bench，简单介绍如何评测的，评测了哪些方面。
We evaluate the performance across a wide range of benchmarks. For hallucination evaluation, we consider
(1) the CHAIR metric \cite{rohrbach2018object}, which evaluates object hallucinations in open-ended caption generation;
(2) POPE \cite{li2023evaluating}, a discriminative benchmark that evaluates object existence consistency via \textit{yes/no} questions;
(3) AMBER \cite{wang2023llm}, a systematic benchmark designed to assess fine-grained hallucinations;
(4) GPT-4 assisted Evaluation \cite{zhao2023beyond} that uses GPT-4 series models to compute Sentence-level Hallucination Ratio (SHR) to detect fine-grained hallucinations.
% , where we utilize the mini version of GPT-4o as it achieves comparable performance for reference-based consistency checking while significantly lower cost, to
In addition, we evaluate the models on two widely used general-purpose benchmarks to assess their overall multimodal capabilities:
(1) MME \cite{fu2025mme}, a comprehensive benchmark measuring multiple dimensions of multimodal proficiency; and
(2) MMBench \cite{liu2024mmbench}, a multiple-choice benchmark analyzing complex multimodal comprehension and reasoning, covering a wide range of perception and cognition tasks.

\textbf{Baselines.}
% 说没有现成的MLRM幻觉缓解Baseline，我们移植了两个MLLM的SOTA基于训练的方法作为基线来比较，来证实我们算法的有效性、
To the best of our knowledge, no prior work has specifically addressed hallucination mitigation for MLRMs. Therefore, we adapt two state-of-the-art (SOTA) training-based methods originally designed for MLLMs as baselines, including RLAIF-V \cite{yu2025rlaif} with iterative DPO and inference-time scaling and the powerful OPA-DPO \cite{yang2025mitigating} that employs on-policy DPO to ensure the alignment of the reference model with the preference data distribution
% an advanced strategy that ensures on-policy consistency by aligning the reference model with the preference data distribution.
% , a representative framework utilizing high-quality AI-generated feedback for robust preference learning

\textbf{Implementation Details.}
% 超参设置，L_{Anc}遵循之前的，设置为1；为L_{DPO}根据不同模型进行消融实验，设置不同的超参，消融实验参考附录X。裁剪的超参设置、SFT的超参设置、DPO阶段的超参设置（如epoch，bs、beta、lr）等。然后遵循先前的研究，我们采用greedy以确保评估的稳定性和可复现性。
% the token retention ratio $\gamma$ is set to 0.9.
In the first SFT stage, we employ Low-Rank Adaptation (LoRA) with a rank of $r=8$ and a scaling factor of $\alpha=16$ for 2 epochs with a learning rate of $5\mathrm{e}^{-5}$ and a global batch size $B=256$. For the subsequent DPO stage, we adopt a learning rate of $1\mathrm{e}^{-6}$ and $B=32$ for 2 epochs of training, with the KL penalty coefficient $\beta=0.1$. As suggested in \cite{wang2024mdpo, yang2025mitigating}, we set $\delta=0$ in Eq.~\eqref{eq:anchor_loss} and $\lambda_{\mathrm{Anc}}=1.0$ for the anchor loss $\mathcal{L}_{\textrm{Anc}}$ in Eq.~\eqref{eq:total_loss}. For the hyperparameter $\lambda_{\mathrm{DPO}}$, we conduct experiments to obtain the optimal value for each model in Appendix \ref{app:add_exps}. Following \cite{yang2025mitigating, yu2025rlaif}, we adopt the greedy decoding for evaluation reproducibility. More details are in Appendix~\ref{app:exp_details}.

\subsection{Performance Evaluation}

\begin{table*}[!t] 
\centering 
\setlength{\tabcolsep}{6.4pt} 
\caption{Comparison of the proposed C3PO with competitive baselines on the CHAIR benchmark. We evaluate on images from MSCOCO. $C_{S}$ and $C_I$ denotes CHAIR$_S$ and CHAIR$_{I}$ respectively.} 
\label{tab:objhal} 
\resizebox{0.9\textwidth}{!}{\begin{tabular}{l|cc|cc|cc|cc|cc} \toprule \multirow{2}{*}{Method} & \multicolumn{2}{c|}{\textbf{R1-Onevision}} & \multicolumn{2}{c|}{\textbf{Orsta-R1}} & \multicolumn{2}{c|}{\textbf{MM-Eureka}} & \multicolumn{2}{c|}{\textbf{MM-R1}} & \multicolumn{2}{c}{\textbf{ThinkLite}} \\ 
& $C_S$ $\downarrow$ & $C_I$ $\downarrow$ & $C_S$ $\downarrow$ & $C_I$ $\downarrow$ & $C_S$ $\downarrow$ & $C_I$ $\downarrow$ & $C_S$ $\downarrow$ & $C_I$ $\downarrow$ & $C_S$ $\downarrow$ & $C_I$ $\downarrow$ \\ \midrule 
\textit{Vanilla} & 36.7 & 8.2 & 40.7 & 8.5 & 44.0 & 8.6 & 37.0 & 7.4 & 41.7 & 7.8\\
RLAIF-V & 39.0 & 8.3 & 39.7 & 7.6 & 41.3 & 8.5 & 38.7 & 7.5 & 40.3 & 7.5\\
OPA-DPO & 36.7 & 7.7 & 41.7 & 8.2 & 41.3 & 7.7 & 36.3 & 7.2 & 40.0 & 7.9\\ 
\rowcolor{gray!15} C3PO & \textbf{35.0} & \textbf{7.4} & \textbf{35.3} & \textbf{6.9} & \textbf{40.0} & \textbf{7.7} & \textbf{35.0} & \textbf{7.2} & \textbf{38.0} & \textbf{7.3} \\  \bottomrule 
\end{tabular}} 
\end{table*}

\textbf{CHAIR Evaluation.}
% 说我们稳定有效，然后有些还让下降，我们的没有
Following \cite{yu2025rlaif, yang2025mitigating}, we instruct the MLRMs to generate captions on 300 images from the MSCOCO validation set \cite{lin2014microsoft} and report results in Table \ref{tab:objhal}. By effectively enhancing the model's reasoning quality, the proposed framework consistently reduces hallucination across all MLRMs, significantly outperforming competitive baselines. \textit{E.g.}, C3PO achieves a 13.27\% and 18.82\% relative reduction of $C_S$ and $C_I$ on Orsta-R1, respectively.

Note that in contrast to consistent hallucination mitigation achieved by our method, baseline methods may even exacerbate hallucinations in certain cases, \textit{e.g.}, RLAIF-V on R1-Onevision and OPA-DPO on Orsta-R1. This phenomenon reveals the instability of transferring methods for MLLMs to the MLRM setting, further highlighting the necessity and effectiveness of the proposed framework, which explicitly accounts for their unique reasoning process.

\begin{table*}[!t]
\centering
\setlength{\tabcolsep}{6pt}
\caption{Comparison of the proposed C3PO with competitive baselines on the POPE benchmark (\%). We report the accuracy and F1 score for evaluation.}
\label{tab:pope}
\resizebox{0.87\textwidth}{!}{\begin{tabular}{llcccccc} \toprule
\multirow{2}{*}{Model} & \multirow{2}{*}{Method} & \multicolumn{2}{c}{Random} & \multicolumn{2}{c}{Popular} & \multicolumn{2}{c}{Adversarial} \\ \cmidrule(lr){3-4} \cmidrule(lr){5-6} \cmidrule(lr){7-8}
& & Acc. & F1 score & Acc. & F1 score & Acc. & F1 score \\ \midrule
% \multirow{4}{*}{Orsta-R1} & \textit{Vanilla} & 85.15 & 83.29 & 84.70 & 82.43 & 84.13 & 81.90 \\
% & RLAIF-V      & 85.95 & 84.32 & 85.37 & 83.36 & 84.57 & 82.61 \\
% & OPA-DPO    & 86.36 & 84.88 & 85.80 & 83.95 & 84.90 & 83.10 \\
% &  \gc C3PO  & \gc\textbf{86.98} &\gc \textbf{85.67} & \gc\textbf{86.27} & \gc\textbf{84.62} & \gc\textbf{85.03} & \gc\textbf{83.45} \\ \midrule
\multirow{4}{*}{MM-Eureka} & \textit{Vanilla} & 84.12 & 81.90 & 83.90 & 81.23 & 83.47 & 80.85 \\
& RLAIF-V      & 83.81 & 81.46 & 83.70 & 80.89 & 83.27 & 80.47 \\
& OPA-DPO    & 85.29 & 83.46 & 84.93 & 82.70 & 84.37 & 82.16 \\
& \gc C3PO  & \gc\textbf{86.22} &\gc \textbf{84.68} &\gc \textbf{85.87} & \gc\textbf{83.94} & \gc\textbf{85.23} & \gc\textbf{83.34} \\ \midrule
\multirow{4}{*}{MM-R1} & \textit{Vanilla} & 84.85 & 82.83 & 84.47 & 82.04 & 83.97 & 81.56 \\
& RLAIF-V      & 85.12 & 83.22 & 84.70 & 82.39 & 84.23 & 81.95 \\
& OPA-DPO    & 85.33 & 83.52 & 85.00 & 82.79 & 84.50 & 82.31 \\
& \gc C3PO &\gc \textbf{86.91} &\gc \textbf{85.57} &\gc \textbf{86.37} & \gc\textbf{84.68} & \gc\textbf{85.57} & \gc\textbf{83.91} \\ \midrule 
\multirow{4}{*}{ThinkLite} & \textit{Vanilla} & 83.23 & 80.63 & 83.03 & 79.97 & 82.77 & 79.72 \\
& RLAIF-V   & 83.71 & 81.29 & 83.37 & 80.50 & 83.00 & 80.14 \\
& OPA-DPO  & 85.02 & 83.07 & 84.70 & 82.34 & 84.10 & 81.77 \\
& \gc C3PO &\gc \textbf{85.98} & \gc\textbf{84.37} &\gc \textbf{85.63} &\gc \textbf{83.63} & \gc\textbf{84.97} & \gc\textbf{83.00} \\ \bottomrule
\end{tabular}}
\end{table*}
\begin{figure*}[!t]
\begin{center}
\includegraphics[width=0.98\linewidth]{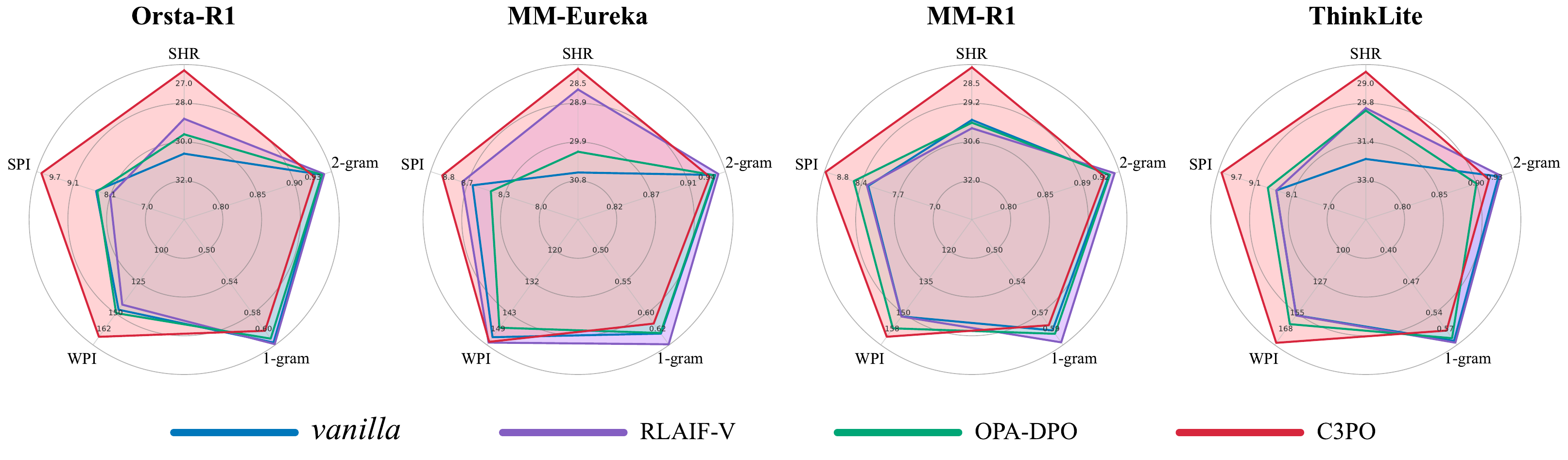}
\end{center}
\caption{GPT-4 assisted benchmark. Sentence-level Hallucination Ratio (SHR) measures the hallucination degree. We also provide 1\&2-gram, the number of sentences per image (SPI), and words per image (WPI). A larger radar area indicates better performance.}
\label{fig:shr}
\end{figure*}

\textbf{POPE Evaluation.}
POPE is another popular evaluation framework for object hallucination, which queries models with “\texttt{Is there a <object> in the image?}” to answer a \textit{yes/no} question. These questions are further categorized into Random, Popular, and Adversarial based on the object type (see Appendix \ref{appendix:benchmarks_intro} for details). We report the accuracy and F1 scores over 3,000 samples in Table~\ref{tab:pope}. The results indicate that the proposed method facilitates more effective reasoning and enables better perception of the images, substantially reducing hallucination rates across all three question types. Due to space limits, results for R1-Onevision and Orsta-R1 are provided in Appendix~\ref{app:add_exps}.

\textbf{GPT-4 Assisted Evaluation.}
% 说单词数和那个啥数分别表明细节程度和词汇多样性。
While CHAIR and POPE are two widely used and reliable hallucination benchmarks, they focus exclusively on object hallucination and do not consider more complex positional, relational, or attribute hallucinations. To evaluate these aspects, we adopt the GPT-Assisted Benchmark \cite{zhao2023beyond}, an LLM-as-judge framework that comprehensively evaluates fine-grained hallucinations using the Visual Genome dataset \cite{krishna2017visual}. Specifically, the MLRM first generates a description for a given image, and the GPT 4-series model is instructed to assess the hallucination level of the generated content based on fine-grained human annotations. 
We additionally report 1\&2- gram metrics for word diversity, as well as SPI and WPI to measure the level of detail.

As shown in Figure~\ref{fig:shr}, C3PO achieves strong overall performance across different MLRMs, significantly reducing fine-grained hallucinations. For instance, C3PO reduces the SHR by 14.05\% for Orsta-R1 compared to the \textit{Vanilla} model. Moreover, it substantially improves response detail (higher WPI and SPI), possibly benefiting from reasoning-enhanced optimization that allows a more comprehensive and detailed perception of the visual input.
\begin{table*}[!t]
\centering
\caption{Comparison of the proposed C3PO with competitive baselines on the AMBER benchmark (\%). We report the accuracy and F1 score for evaluation.}
\label{tab:amber}
\setlength{\tabcolsep}{4.7pt}
\resizebox{0.98\textwidth}{!}{
\begin{tabular}{l|cc|cc|cc|cc|cc}
\toprule
\multirow{2}{*}{Method}
& \multicolumn{2}{c|}{\textbf{R1-Onevision}}
& \multicolumn{2}{c|}{\textbf{Orsta-R1}}
& \multicolumn{2}{c|}{\textbf{MM-Eureka}}
& \multicolumn{2}{c|}{\textbf{MM-R1}}
& \multicolumn{2}{c}{\textbf{ThinkLite}}\\
& Acc. & F1 score
& Acc. & F1 score
& Acc. & F1 score
& Acc. & F1 score
& Acc. & F1 score\\
\midrule
\textit{Vanilla} & 81.52 & 86.54 & 82.94 & 87.31 & 83.27 & 87.15 & 84.38 & 88.20 & 82.25 & 86.96\\
RLAIF-V      & 80.99 & 84.97 & 83.14 & 87.50 & 83.46 & 87.49 & 84.17 & 88.11 & 82.31 & 87.04\\
OPA-DPO      & 80.71 & 85.30 & 83.58 & 87.67 & 83.72 & 87.59 & 84.76 & 88.51 & 83.02 & 87.36\\
\rowcolor{gray!15}
C3PO & \textbf{81.90} & \textbf{86.73} & \textbf{84.57} & \textbf{88.50} & \textbf{85.01} & \textbf{88.80} & \textbf{85.61} & \textbf{89.25} & \textbf{83.96} & \textbf{88.26}\\
\bottomrule
\end{tabular}}
\end{table*}

\textbf{AMBER Evaluation.}
AMBER \cite{wang2023llm} is another challenging benchmark designed to measure models' multi-dimensional hallucinations, including existence, attribute, and relation hallucinations. Following \cite{yu2025rlaif}, we focus on the discriminative part for an objective and reproducible evaluation and report accuracy and F1 scores on more than 15,000 carefully designed yes/no questions in Table~\ref{tab:amber}. While baselines provide only limited improvements or even degrade performance, the proposed C3PO consistently achieves improvements in hallucination mitigation, again demonstrating the necessity and effectiveness of the proposed framework.

%%%%%%%%%%%%%%%%%%%%%%%%%%%%%%%%%
\begin{figure}[t]
\centering
\begin{minipage}{0.4905\linewidth}
    \centering
    \includegraphics[width=\linewidth]{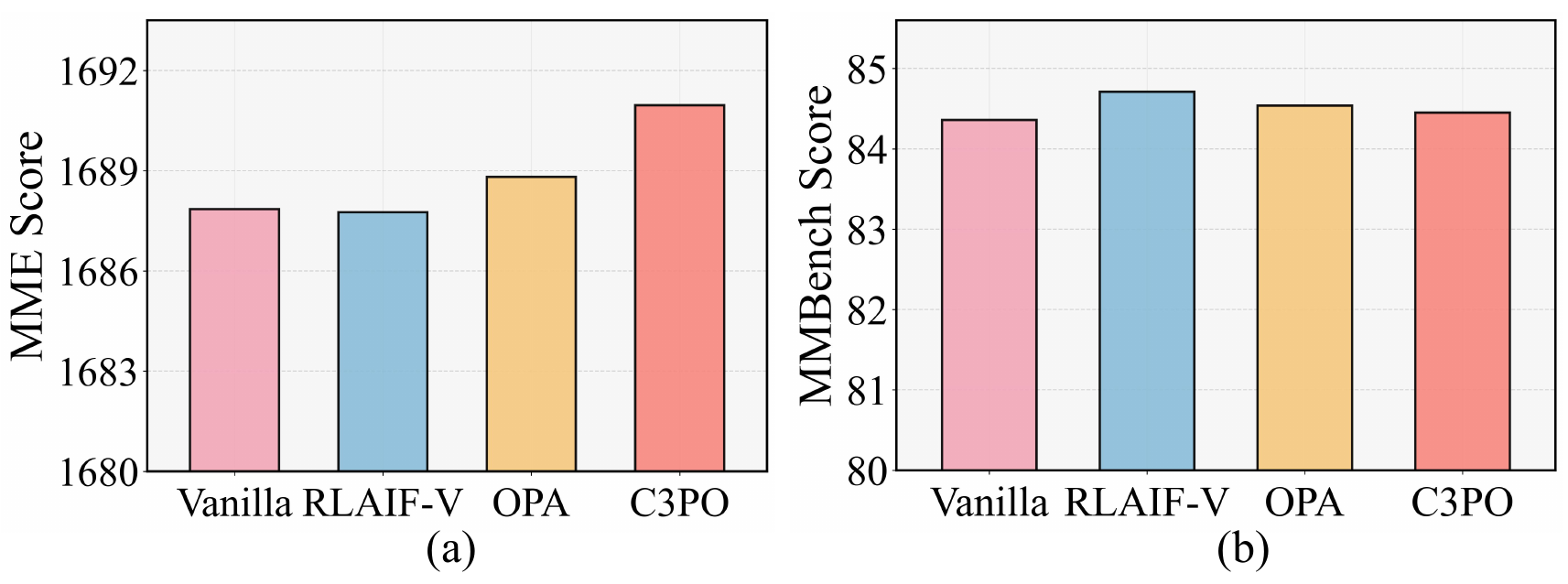}
    \caption{Performance of different methods on two general-purpose benchmarks.}
    \label{fig:mme}
\end{minipage}
\hfill
\begin{minipage}{0.493\linewidth}
    \centering
    \includegraphics[width=\linewidth]{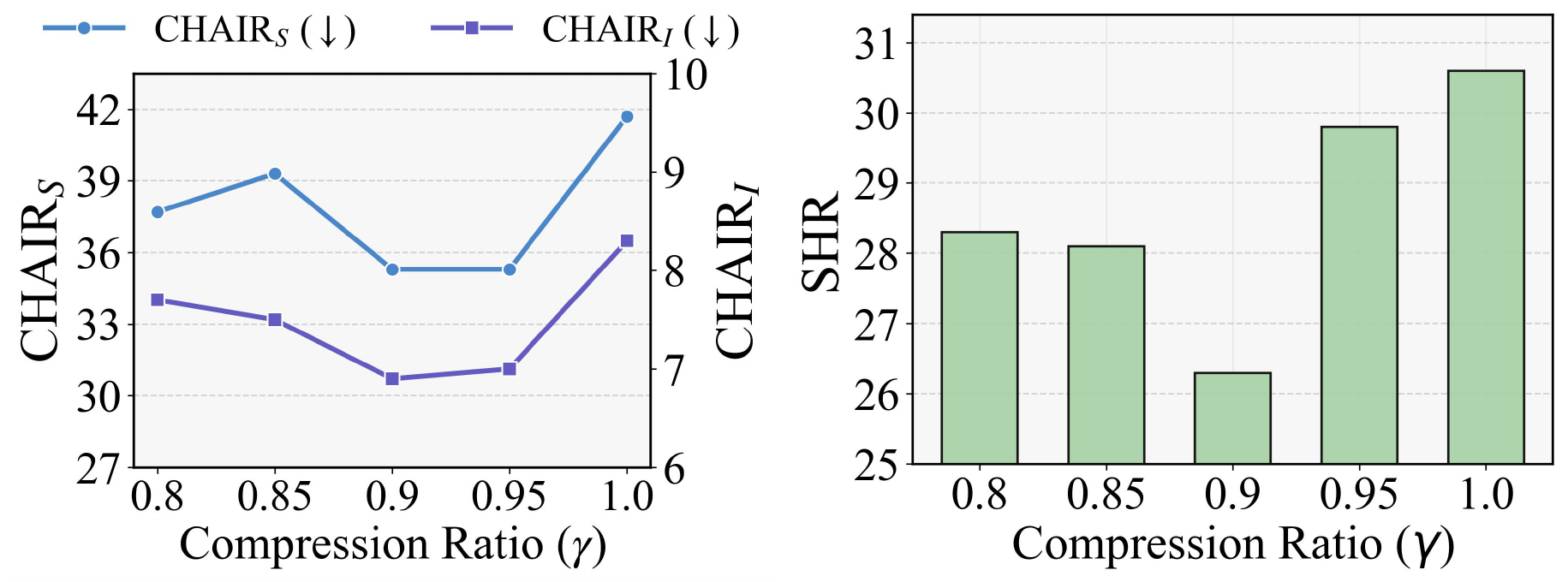}
    \caption{Ablation study of the compression ratio $\gamma$ on CHAIR and SHR.}
    \label{fig:ablation}
\end{minipage}
\end{figure}
%%%%%%%%%%%%%%%%%%%%%%%%%%%%%%%%%

%%%%%%%%%%%%%%%%%%%%%%%%%%%%%%%%%
% \begin{figure}[t]
% \begin{center}
% \includegraphics[width=\linewidth]{figs/MME_MMBench.pdf}
% \end{center}
% % \vspace{-0.8em}
% \caption{Performance on two general-purpose benchmarks.}
% \label{fig:mme}
% \vspace{-0.1em}
% \end{figure}

% \begin{figure}[!t]
% \begin{center}
% \includegraphics[width=\linewidth]{figs/ablation_gamma.pdf}
% \end{center}
% \caption{Ablation study of the compression ratio $\gamma$.}
% \label{fig:ablation}
% \end{figure}
%%%%%%%%%%%%%%%%%%%%%%%%%%%%%

\textbf{MME and MMBench Evaluations.}
In addition to hallucination benchmarks, we additionally consider two popular general-purpose benchmarks that measure general multimodal capabilities. MME provides a suite of fine-grained and multiple-choice questions across various categories. We align with \cite{huang2024opera} and report the overall perception score covering 10 multimodal sub-tasks. MMBench is another large-scale benchmark consisting of over 3,000 curated multiple-choice questions. We compute the average score across 20 complex reasoning tasks.
We provide results on the representative Orsta-R1 model in Figure \ref{fig:mme}. As observed, MRLMs trained with the proposed reasoning-oriented strategy generally preserve their multimodal capabilities and even improve over the base model.

\subsection{Ablation Study}
% 主要分析趋势，裁剪冗余，但保留关键信息。同时放一个为1的，就说加了的显著还是比没加好证实所提出算法有效性。
In this section, we adopt Orsta-R1 as a representative model to investigate the impact of key hyperparameters and quantify the contribution of each proposed technique. 
Note that we provide ablation using more MLRMs in Appendix~\ref{app:add_exps} due to page limits.

\textbf{Ablation on the token compression ratio.} 
During the construction of the SFT data, $\gamma$ is a critical hyperparameter that controls the number of remaining tokens in the reasoning chain. A larger $\gamma$ may fail to sufficiently remove noise and redundancy, while a smaller $\gamma$ may unnecessarily discard informative tokens and thus impair model performance. We provide an ablation study of $\gamma$ in Figure~\ref{fig:ablation}.
As shown, $\gamma=0.9$ achieves optimal performance on both CHAIR and GPT-4 assisted evaluations. Moreover, we observe that pruning rates smaller than $1.0$ generally outperform the no pruning ($\gamma=1$), which again validates the effectiveness of our CoT compression strategy.

% to ensure well-formed and reliable predictions.
\setlength{\intextsep}{2.6pt}
\setlength{\abovecaptionskip}{2.8pt}
\setlength{\belowcaptionskip}{2.6pt}
\begin{wraptable}{l}{0.45\linewidth}
\centering
% \vspace{-1em}
\caption{Ablation study of the proposed techniques in C3PO under CHAIR and GPT-4 assisted evaluation benchmarks.}
\label{tab:ablation_main}
\resizebox{\linewidth}{!}{
\begin{tabular}{l|cc|c}
\toprule
\multirow{2}{*}{Method} & \multicolumn{2}{c|}{CHAIR Eval} & \multicolumn{1}{c}{GPT-4 Eval} \\
& $C_S$ $\downarrow$ & $C_I$ $\downarrow$ & SHR $\downarrow$ \\
\midrule
\rowcolor{gray!15} C3PO & \textbf{35.3} & \textbf{6.9} & \textbf{26.3} \\
w/o loss $\mathcal{L}_{\mathrm{RE}}$ & 37.3 & 7.2 & 28.8 \\
w/o loss $\mathcal{L}_{\mathrm{DPO}}$ & 39.7 & 8.1 & 29.4 \\
w/o loss $\mathcal{L}_{\mathrm{Anc}}$ & 37.7 & 7.5 & 27.2 \\
w/o MHI strategy & 37.0 & 7.9 & 28.4 \\
\bottomrule
\end{tabular}}
% \vspace{-1em}
\end{wraptable}

\textbf{Ablation on the designed techniques.} 
In addition to CoT compression, we then verify the effectiveness of the other proposed techniques. Specifically, we evaluate the designed loss functions $\mathcal{L}_{\mathrm{RE}}$, $\mathcal{L}_{\mathrm{DPO}}$, and $\mathcal{L}_{\mathrm{Anc}}$, as well as the proposed Multimodal Hallucination-Inducing (MHI) strategy. 
% The quantitative results are provided in Table~\ref{tab:ablation_main}.

Table~\ref{tab:ablation_main} confirms the contribution of each component, validating the rationale behind our designs. Notably, removing $\mathcal{L}_{\mathrm{DPO}}$ leads to a significant performance drop, showing the necessity of modeling the entire output distribution as a regularization.

\section{Conclusion}
This work investigates the critical issue of hallucinations in MLRMs. We begin by analyzing two key factors contributing to hallucination emergence, based on which we propose C3PO, a two-stage training-based mitigation framework that first adopts SFT to remove noise and redundancy in generated reasoning trajectories, and then explicitly enhances reasoning via contrastive preference optimization with carefully constructed preference pairs.
We theoretically justify the superiority of the proposed techniques from the information bottleneck principle. Extensive experiments across diverse MLRMs and benchmarks validate its effectiveness and generality. We highlight that this work makes a pioneering step in mitigating hallucinations in MLRMs, deepening the understanding of the underlying mechanisms, and inspiring future studies on mitigation strategies.

\section*{Acknowledgement}
This work is supported in part by the National Natural Science Foundation of China under grant 62301189, 62576122,62571298, Guangdong Basic and Applied Basic Research Foundation under grant 2026A1515011139.

{
\small
\bibliography{example_paper}
\bibliographystyle{unsrt}
}

%%%%%%%%%%%%%%%%%%%%%%%%%%%%%%%%%%%%%%%%%%%%%%%%%%%%%%%%%%%%
\newpage
\appendix

\section{Proofs for Theorems}
\label{sec:证明}
\setcounter{theorem}{0}

Here we prove our theorems in Section~\ref{sec:理论分析}.

\begin{theorem}[Compression reduces IB]
Let the CoT $Z$ be decomposed into $Z_{\mathrm{ret}}$ and $Z_{\mathrm{trim}}$, where $Z_{\mathrm{ret}}$ denotes the retaining task-critical part, $Z_{\mathrm{trim}}$ represents the trimmed redundant part, $Z_{\mathrm{ret}}\cup Z_{\mathrm{trim}}=Z$ and $Z_{\mathrm{ret}}\cap Z_{\mathrm{trim}}=\emptyset$. Then, the IB objective satisfies
\begin{equation}
\mathcal{L}_\textnormal{IB}(Z)\ge\mathcal{L}_\textnormal{IB}(Z_{\mathrm{ret}}).
\end{equation}
\end{theorem}

\begin{proof}

According to $Z=Z_{\mathrm{ret}}\cup Z_{\mathrm{trim}}$ and the chain rule of mutual information, we have
\begin{equation}
\label{eq:互信息链式法则}
\begin{alignedat}{3}
&\mathcal{I}(X;Z)&&=\mathcal{I}(X;Z_\text{ret})&&+\mathcal{I}(X;Z_\text{trim}|Z_\text{ret}),\\
&\mathcal{I}(Y;Z)&&=\mathcal{I}(Y;Z_\text{ret})&&+\mathcal{I}(Y;Z_\text{trim}|Z_\text{ret}).
\end{alignedat}
\end{equation}

When $Z_\text{ret}$ is given, $Z_\text{trim}$ can be relevant to the input $X$ but irrelevant to the ground truth $Y$, because our compressor only trims low-importance words without contributions to final predictions (see Appendix \ref{app:visual_pruned_cot}). Therefore
\begin{equation}
\label{eq:条件互信息}
\mathcal{I}(X;Z_\text{trim}|Z_\text{ret})\ge0,\quad
\mathcal{I}(Y;Z_\text{trim}|Z_\text{ret})=0.
\end{equation}

\noindent(\ref{eq:互信息链式法则}) and (\ref{eq:条件互信息}) yield
\begin{equation}
\mathcal{I}(X;Z)\ge\mathcal{I}(X;Z_\text{ret}),\quad
\mathcal{I}(Y;Z)=\mathcal{I}(Y;Z_\text{ret}).
\end{equation}

Then we conclude that
\begin{equation}
\begin{aligned}
\mathcal{L}_\textnormal{IB}(Z)
&=\mathcal{I}(X;Z)-\lambda_\text{IB}\mathcal{I}(Y;Z),\\
&\ge\mathcal{I}(X;Z_\text{ret})-\lambda_\text{IB}\mathcal{I}(Y;Z_\text{ret}),\\
&=\mathcal{L}_\textnormal{IB}(Z_\text{ret}).
\end{aligned}
\end{equation}

\end{proof}

\begin{theorem}[Enhancement reduces IB]
Let the CoT $Z$ be refined to $Z'$ via our reasoning-enhanced optimization, which supplements information about the ground truth $Y$.
Then, the IB objective satisfies
\begin{equation}
\mathcal{L}_\textnormal{IB}(Z) > \mathcal{L}_\textnormal{IB}(Z').
\end{equation}
\end{theorem}

\begin{proof}

Note that the correct answer $Y$ is contained within the input $X$, which implies
\begin{equation}
\mathcal{I}(X;Z)=\mathcal{I}(X,Y;Z),\quad
\mathcal{I}(X;Z')=\mathcal{I}(X,Y;Z').
\end{equation}

In addition, according to the chain rule of mutual information, we have
\begin{equation}
\begin{alignedat}{3}
&\mathcal{I}(X;Z) &&=\mathcal{I}(Y;Z) &&+\mathcal{I}(X;Z|Y),\\
&\mathcal{I}(X;Z')&&=\mathcal{I}(Y;Z')&&+\mathcal{I}(X;Z'|Y).
\end{alignedat}
\end{equation}

Our reasoning-enhanced optimization guides the CoT towards the correct answer, thereby reducing the uncertainty of $Y$ when $Z'$ is given. Formally, it makes
\begin{equation}
\mathcal{H}(Y|Z)>\mathcal{H}(Y|Z'),
\end{equation}
where $\mathcal{H}$ is Shannon entropy to quantify uncertainty.
Then
\begin{equation}
\begin{aligned}
\mathcal{I}(Y;Z)&=\mathcal{H}(Y)-\mathcal{H}(Y|Z),\\
&<\mathcal{H}(Y)-\mathcal{H}(Y|Z'),\\
&=\mathcal{I}(Y;Z').
\end{aligned}
\end{equation}

The information supplemented to $Z'$ by the reasoning-enhanced optimization can essentially be contained within the ground truth $Y$, which implies that
\begin{equation}
\mathcal{H}(Z'|Z,Y)=0.
\end{equation}

Then we have
\begin{equation}
\begin{aligned}
\mathcal{I}(X;Z'|Y)&\le\mathcal{I}(X;Z,Z'|Y),\\
&=\mathcal{I}(X;Z|Y)+\mathcal{I}(X;Z'|Z,Y),\\
&\le\mathcal{I}(X;Z|Y)+\mathcal{H}(Z'|Z,Y),\\
&=\mathcal{I}(X;Z|Y).
\end{aligned}
\end{equation}
Recall that we focus on $\lambda_{\textrm{IB}}>1$ to prioritize accuracy over compression. Then we conclude that
\begin{equation}
\begin{aligned}
\mathcal{L}_\textnormal{IB}(Z)
&=\mathcal{I}(X;Z)-\lambda_\text{IB}\mathcal{I}(Y;Z),\\
&=\mathcal{I}(Y;Z)+\mathcal{I}(X;Z|Y)-\lambda_\text{IB}\mathcal{I}(Y;Z),\\
&=\mathcal{I}(X;Z|Y)-(\lambda_\text{IB}-1)\mathcal{I}(Y;Z),\\
&>\mathcal{I}(X;Z'|Y)-(\lambda_\text{IB}-1)\mathcal{I}(Y;Z'),\\
&=\mathcal{I}(Y;Z')+\mathcal{I}(X;Z'|Y)-\lambda_\text{IB}\mathcal{I}(Y;Z'),\\
&=\mathcal{I}(X;Z')-\lambda_\text{IB}\mathcal{I}(Y;Z'),\\
&=\mathcal{L}_\textnormal{IB}(Z').
\end{aligned}
\end{equation}
\end{proof}

\section{Experimental Details}
% 可以把benchmark的介绍详细展开？
\label{app:exp_details}
This part provides more experimental details regarding data construction and evaluation benchmarks. 
Note that all the experiments are performed with four 48G NVIDIA A6000 GPUs.
\subsection{Data Construction}
\label{appendix:data_construction}
We randomly sample 20k image-question pairs from the RLAIF-V \cite{yu2025rlaif} dataset. Each instance provides an image $v$, a question $x$, and a reference answer $y_{\mathrm{GT}}$.

\textbf{SFT Dataset Construction.} For each instance $(v,x)$, we query the target model to generate a structured response that contains an explicit CoT and a final answer. The output is parsed into an original reasoning trace $z_{\mathrm{Gen}}$ and an answer $y_{\mathrm{Gen}}$ based on the \texttt{<think>} and \texttt{<answer>} tags. We apply LLMlingua-2 to filter redundant tokens in reasoning traces. By setting the compression ratio to $\gamma=0.9$, we obtain the compressed reasoning chain $z'=\mathcal{C}(z;\gamma)$. Our SFT data is constructed directly from the compressed reasoning trace paired with the model's original answer. 
% Concretely, each target in the SFT Dataset ($\mathcal{D}_s$) takes the form $z'_\mathrm{Gen}\,\Vert\,y_\mathrm{Gen}$. 
This design reduces redundant textual tokens in CoT, increasing the relative salience of visually relevant evidence during training. We train a reference model using this dataset via SFT, which subsequently serves as the generator for all negative samples in the preference learning stage.

\textbf{Preference Dataset Construction.} To construct the preference dataset, we employ a stronger evaluator Qwen3-VL-30B-Instruct to minimally refine the original reasoning trace and subsequently apply LLMlingua-2 to compress it, yielding the final corrected reasoning chain $z_{\mathrm{Rev}}$. The goal of this step is to correct factual inconsistencies with image content while remaining faithful to the model's stylistic characteristics and reasoning structure. We follow the revision prompt designed in \cite{yang2025mitigating}, the key refinement instruction is presented as follows:
\begin{tcolorbox}[title=Refinement Prompt:]
Your role is as a discerning assistant tasked with evaluating and refining responses for multimodal tasks. Upon being presented with a question that requires the interpretation of both text and images, you will receive two distinct responses. The first is crafted by our sophisticated multimodal model, while the second represents an approximate ideal answer—it may be incomplete or incorrect. You will also be provided with the images pertinent to the question. Your objective is to meticulously assess these responses. You are to enhance the model-generated response by making precise, minimal modifications that bring it into closer alignment with both the image and the approximate ideal answer. Your revisions should preserve the integrity of the original response as much as possible.
\end{tcolorbox}
We then construct the positive sample by concatenating this refined chain with the ground truth answer: $z_{\mathrm{Rev}}\,\Vert\,y_{\mathrm{GT}}$. This ensures that the positive sample acts as a gold standard, being both factually accurate regarding the visual content and structurally optimized for concise and efficient reasoning. For each training instance $(v,x)$, we pair the positive sample with three distinct negative samples derived from the reference model. These negatives comprise: (i) a \textit{natural negative}, generated directly by the reference model given the original image and question, capturing the model's intrinsic tendency to generate non-factual content; (ii) a \textit{visually induced negative}, elicited by applying a random masking ratio of $0.3$ to the visual tokens, which forces the model to hallucinate based on language priors due to incomplete visual information; (iii) an \textit{instruction-induced negative}, crafted by appending a specific system prompt that explicitly steers the model towards fabrication: 
\begin{tcolorbox}[title=Misleading Prompt:]
Generate a response to the question about the image. Focus on using your language prior to creating a plausible-sounding reasoning trace and answer, but intentionally introduce hallucinations—details that seem related to the image but are not actually supported by it. Ignore the image content when necessary to make the response imaginative yet coherent.
\end{tcolorbox}
% The resulting Preference Dataset $\mathcal{D}_p$ consists of these three categories of positive-negative pairs.

\subsection{Evaluation Benchmarks}
\label{appendix:benchmarks_intro}
\textbf{Object HalBench Evaluations.} To evaluate object-level hallucinations in open-ended image description, the CHAIR (Caption Hallucination Assessment with Image Relevance) metric is introduced. This rigorous metric quantifies the discrepancy between objects generated in captions and those present in the ground-truth annotations. Both instance-level ($\mathrm{CHAIR}_I$) and sentence-level ($\mathrm{CHAIR}_S$) scores are reported, where lower values indicate a higher degree of factual consistency. The mathematical formulations for these two variants are defined as follows:

\begin{align*}
C_I &= \frac{|\text{hallucinated objects}|}{|\text{all mentioned objects}|} \\[10pt]
C_S &= \frac{|\text{captions with hallucinated objects}|}{|\text{all captions}|}
\end{align*}

% 这里看看要不要加CHAIR的公式
\textbf{AMBER Evaluations.}
AMBER establishes a versatile framework for evaluating hallucinations in generative and discriminative scenarios. Following \cite{yu2025rlaif}, we consider the discriminative tasks to assess fine-grained visual grounding for objective and reproducible evaluation. This benchmark formulates hallucination assessment as a rigorous binary classification problem, probing the model with specific ``Yes/No'' queries to verify visual facts. It comprehensively covers three dimensions of hallucination: (i) \textit{Existence}, verifying the presence of specific objects; (ii) \textit{Attribute}, evaluating detailed properties including object states, counts, and actions; and (iii) \textit{Relation}, assessing spatial relationships and interactions between objects. For R1-Onevision responses that do not begin with Yes/No as instructed, we use Qwen3-VL-30B-Instruct to extract the binary answer. We report both Accuracy and F1 scores to quantify the model's ability to distinguish truthful details from fabrications.

\textbf{GPT-4 Assisted Evaluations.} To quantify the degree of hallucination in detailed image descriptions, the Sentence-level Hallucination Ratio (SHR) is adopted. Following established protocols, models are prompted to describe images from the Visual Genome dataset \cite{krishna2017visual} in detail. The mini version of GPT-4o is utilized as a cost-effective and capable evaluator to identify contradictory statements in the generated text compared to the ground-truth annotations provided by the Visual Genome.

\textbf{POPE Evaluations.} To eliminate the ambiguity often found in free-form generation metrics, the Polling-based Object Probing Evaluation (POPE) reformulates hallucination assessment as a strict binary verification task. An equal distribution of positive and negative samples is enforced to eliminate the influence of prior label frequency. The evaluation includes three distinct difficulty levels: (i) \textit{random} sampling serves as a standard baseline; (ii) \textit{popular} sampling utilizes frequent categories to test resistance against popularity bias; and (iii) \textit{adversarial} sampling employs co-occurring objects to challenge fine-grained visual discrimination. We report Accuracy and F1 scores to evaluate the reliability of object verification across these diverse settings.

\textbf{MME and MMBench Evaluations.} To ensure that the hallucination mitigation strategy does not compromise general multimodal capabilities, we conduct evaluations on two comprehensive benchmarks. MME is employed to assess a wide spectrum of perception and cognition tasks. By employing a suite of binary-choice questions with a consistent answer format, MME facilitates a precise measurement of a model's ability to remain faithful to visual content. MMBench is utilized to evaluate complex multimodal comprehension across 20 distinct ability dimensions ranging from basic perception to high-level reasoning. To ensure robustness, it introduces the CircularEval strategy, which requires consistent predictions across multiple permutations of choices to eliminate the influence of random guessing. Together, these evaluations provide an effective validation of the model's ability to preserve general multimodal understanding and instruction-following proficiency.

\begin{table}[!t]
\centering
\caption{Additional performance comparison of the proposed C3PO and strong baselines on the POPE benchmark. We supplement results on MLRMs Orsta-R1 and R1-Onevision.}
\label{tab:pope_addtional}
\resizebox{0.9\textwidth}{!}{\begin{tabular}{llcccccc} \toprule
\multirow{2}{*}{Model} & \multirow{2}{*}{Method} & \multicolumn{2}{c}{Random} & \multicolumn{2}{c}{Popular} & \multicolumn{2}{c}{Adversarial} \\ \cmidrule(lr){3-4} \cmidrule(lr){5-6} \cmidrule(lr){7-8}
& & Accuracy & F1 score & Accuracy & F1 score & Accuracy & F1 score \\ \midrule
\multirow{4}{*}{R1-Onevision} & \textit{Vanilla} & 83.64 & 81.82 & \textbf{82.80} & 80.59 & 81.93 & 79.81 \\
& RLAIF-V   & 82.85 & 81.03 & 81.60 & 79.45 & 81.57 & 79.42 \\
& OPA-DPO  & 83.85 & 82.28 & 82.57 & 80.67 & 82.03 & 80.19 \\
& \gc C3PO &\gc \textbf{83.88} & \gc\textbf{82.66} &\gc 82.53 &\gc \textbf{81.01} & \gc\textbf{82.37} & \gc\textbf{80.87} \\ \bottomrule
\multirow{4}{*}{Orsta-R1} & \textit{Vanilla} & 85.15 & 83.29 & 84.70 & 82.43 & 84.13 & 81.90 \\
& RLAIF-V      & 85.95 & 84.32 & 85.37 & 83.36 & 84.57 & 82.61 \\
& OPA-DPO    & 86.36 & 84.88 & 85.80 & 83.95 & 84.90 & 83.10 \\
&  \gc C3PO  & \gc\textbf{86.98} &\gc \textbf{85.67} & \gc\textbf{86.27} & \gc\textbf{84.62} & \gc\textbf{85.03} & \gc\textbf{83.45} \\ \midrule
\end{tabular}}
\end{table}
\section{Additional Results}
% 折线图做λ_DPO的消融、R1雷达图，R1和Orsta的POPE，以及只加tokenskip的消融。
\label{app:add_exps}

\textbf{POPE evaluation on more MLRMs.}
As a supplement, we report POPE results on additional MLRMs, including R1-Onevision and Orsta-R1. The quantitative results in Table~\ref{tab:pope_addtional} further corroborate the effectiveness of our method in mitigating object hallucinations, outperforming baseline methods. This validates the necessity and efficacy of our CoT compression and preference optimization.

\begin{wrapfigure}{l}{0.45\textwidth}
\begin{center}
\includegraphics[width=\linewidth]{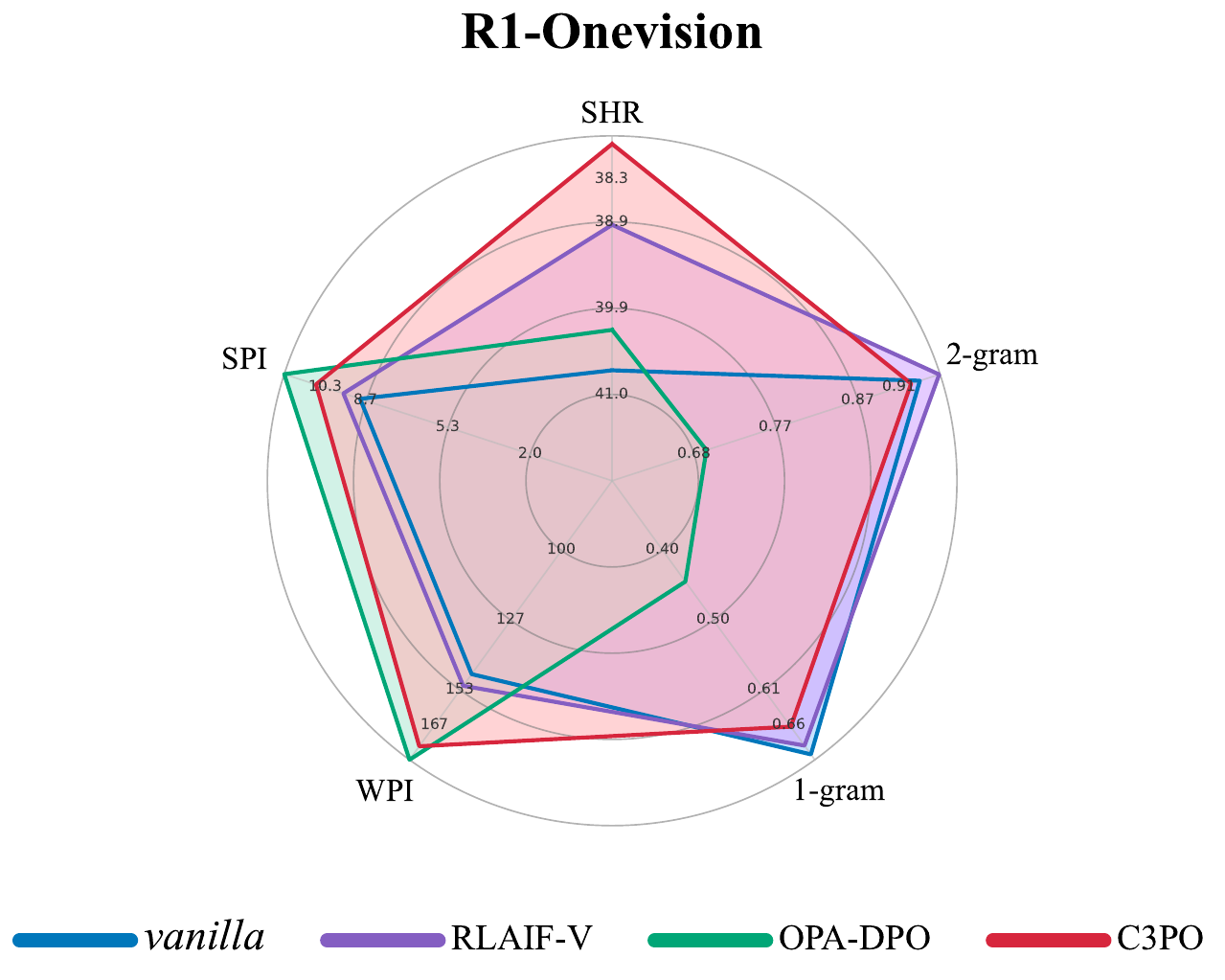}
\end{center}
\caption{GPT-4 assisted benchmark results for R1-Onevision. A larger radar area indicates better general performance.}
\label{fig:r1_radar}
\end{wrapfigure}

\textbf{GPT-4 assisted evaluation on R1-Onevision.}
We further supplement the GPT-4 assisted benchmark results on R1-OneVision. As illustrated in Figure~\ref{fig:r1_radar}, the proposed C3PO significantly outperforms existing SOTA baselines. For instance, in comparison with competitive baseline OPA-DPO, C3PO attains a relative improvement of 5.7\% on the SHR hallucination metric while maintaining comparable text fluency and details. The results again indicate that the proposed redundant CoT token compression, combined with reasoning-enhanced contrastive preference optimization, synergistically suppresses hallucinations without compromising generation quality.

\begin{table}[t]
\centering
\caption{Ablation study of C3PO on the preference optimization and the CoT compression.}
\label{tab:ablation_tokenskip_only}
\resizebox{\linewidth}{!}{
\setlength{\tabcolsep}{1pt}
\begin{tabular}{l|ccc|ccc|ccc|ccc|ccc}
\toprule
\multirow{2}{*}{Metric} & \multicolumn{3}{c|}{R1-Onevision} & \multicolumn{3}{c|}{Orsta-R1} & \multicolumn{3}{c|}{MM-Eureka} & \multicolumn{3}{c|}{MM-R1} & \multicolumn{3}{c}{ThinkLite} \\
& \cellcolor{gray!15} C3PO & w/o DPO & \textit{Vanilla} & \cellcolor{gray!15} C3PO & w/o DPO & \textit{Vanilla} & \cellcolor{gray!15} C3PO & w/o DPO & \textit{Vanilla} & \cellcolor{gray!15} C3PO & w/o DPO & \textit{Vanilla} & \cellcolor{gray!15} C3PO & w/o DPO & \textit{Vanilla} \\
\midrule
$C_S$ $\downarrow$ & \cellcolor{gray!15}\textbf{35.0} & 40.0 & 36.7 & \cellcolor{gray!15}\textbf{35.3} & 39.7 & 40.7 & \cellcolor{gray!15}\textbf{40.0} & 40.7 & 44.0 & \cellcolor{gray!15}\textbf{35.0} & 35.3 & 37.0 & \cellcolor{gray!15}\textbf{38.0} & 41.3 & 41.7 \\
$C_I$ $\downarrow$ & \cellcolor{gray!15}\textbf{7.4} & 7.7 & 8.2 & \cellcolor{gray!15}\textbf{6.9} & 7.7 & 8.5 & \cellcolor{gray!15}\textbf{7.7} & 8.3 & 8.6 & \cellcolor{gray!15} 7.2 & \textbf{6.9} & 7.4 & \cellcolor{gray!15}\textbf{7.3} & 7.2 & 7.8 \\
\midrule
SHR $\downarrow$ & \cellcolor{gray!15} 37.9 & \textbf{36.7} & 40.7 & \cellcolor{gray!15}\textbf{26.3} & 29.2 & 30.6 & \cellcolor{gray!15}\textbf{28.1} & 29.3 & 30.6 & \cellcolor{gray!15}\textbf{27.9} & 31.0 & 29.8 & \cellcolor{gray!15}\textbf{28.5} & 29.9 & 32.1 \\
\bottomrule
\end{tabular}}
\end{table}

\textbf{Ablation on preference optimization and SFT.}
To evaluate the effectiveness of the preference optimization, we design a variant that undergoes only SFT on compressed text, without subsequent preference optimization. As demonstrated in Table~\ref{tab:ablation_tokenskip_only}, supervised fine-tuning on compressed text already leads to notable improvements across both evaluation benchmarks. Building upon this foundation, C3PO further achieves reductions of $4.4$ and $2.9$ in $C_S$ and SHR when equipped with the proposed powerful preference optimization. Overall, these results indicate that the synergy between CoT token compression and DPO allows C3PO to achieve stronger performance gains.
% \begin{figure}[!t]
% \centering
% \includegraphics[width=0.48\linewidth]{figs/lambda_dpo_loss_R1-Onevision.pdf}
% \vspace{0.8em}
% \includegraphics[width=0.48\linewidth]{figs/lambda_dpo_loss_Orsta-R1.pdf}
% \vspace{0.8em}
% \includegraphics[width=0.48\linewidth]{figs/lambda_dpo_loss_MM-Eureka.pdf}
% \vspace{0.8em}
% \includegraphics[width=0.48\linewidth]{figs/lambda_dpo_loss_MM-R1.pdf}
% \vspace{0.8em}
% \includegraphics[width=0.48\linewidth]{figs/lambda_dpo_loss_ThinkLite.pdf}
% \caption{
% Ablation study on the weight factor $\lambda_{\mathrm{DPO}}$ of the $\mathcal{L}_{\mathrm{DPO}}$ term across different MRLMs.}
% \label{fig:lambda_dpo_loss_all}
% \vspace{-0.5em}
% \end{figure}

\begin{wrapfigure}{r}{0.48\textwidth}
    \centering
    \vspace{-0.8em}
    \includegraphics[width=\linewidth]{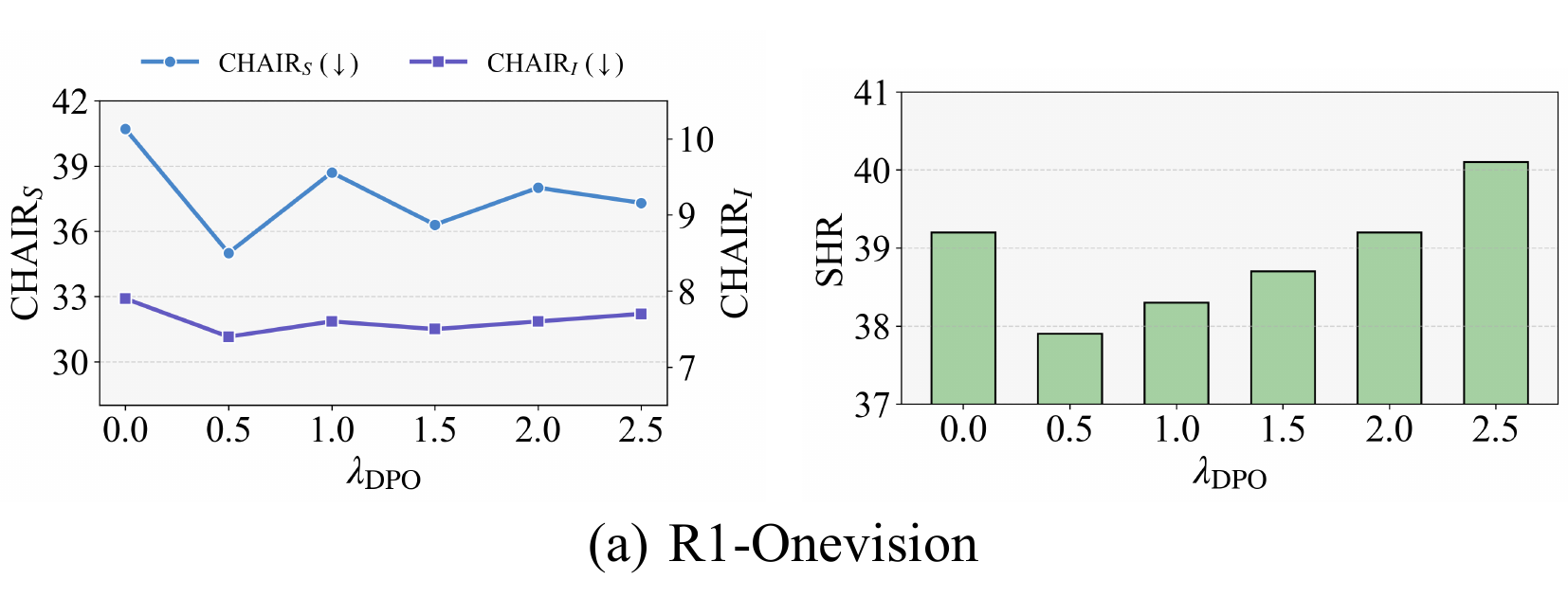}
    \vspace{0.5em}
    \includegraphics[width=\linewidth]{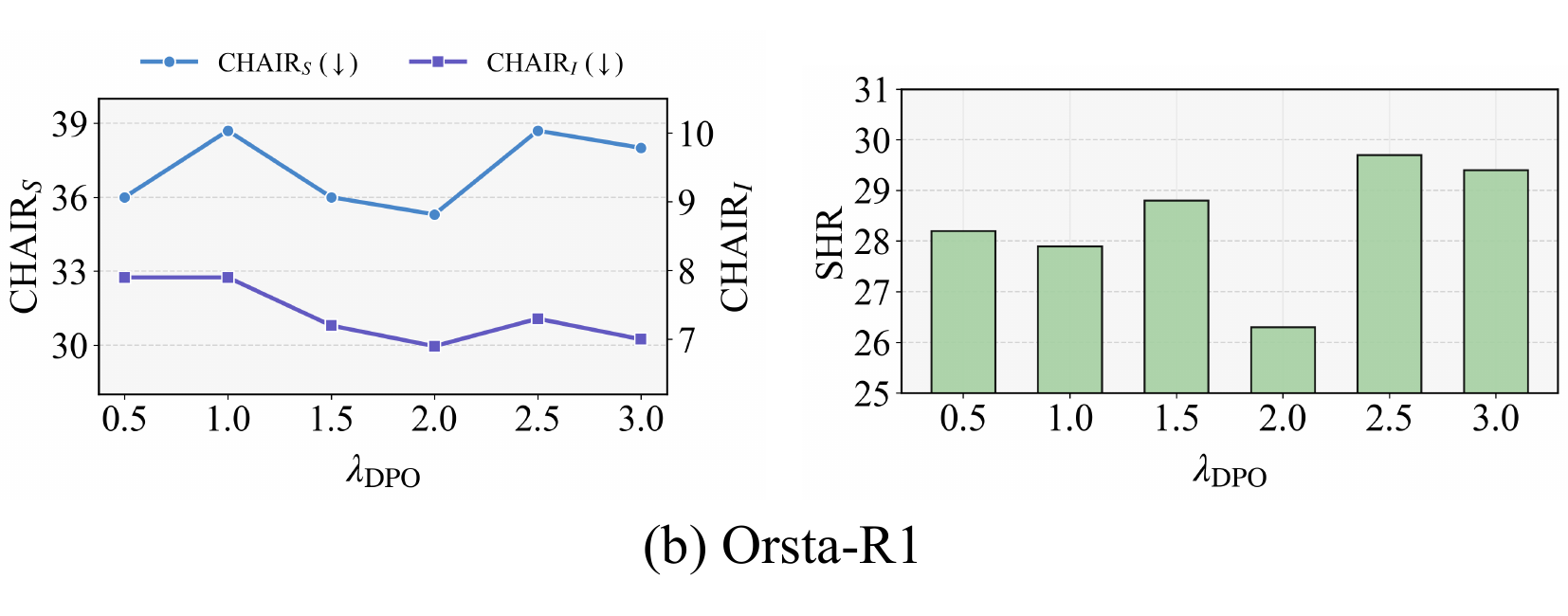}
    \vspace{0.5em}
    \includegraphics[width=\linewidth]{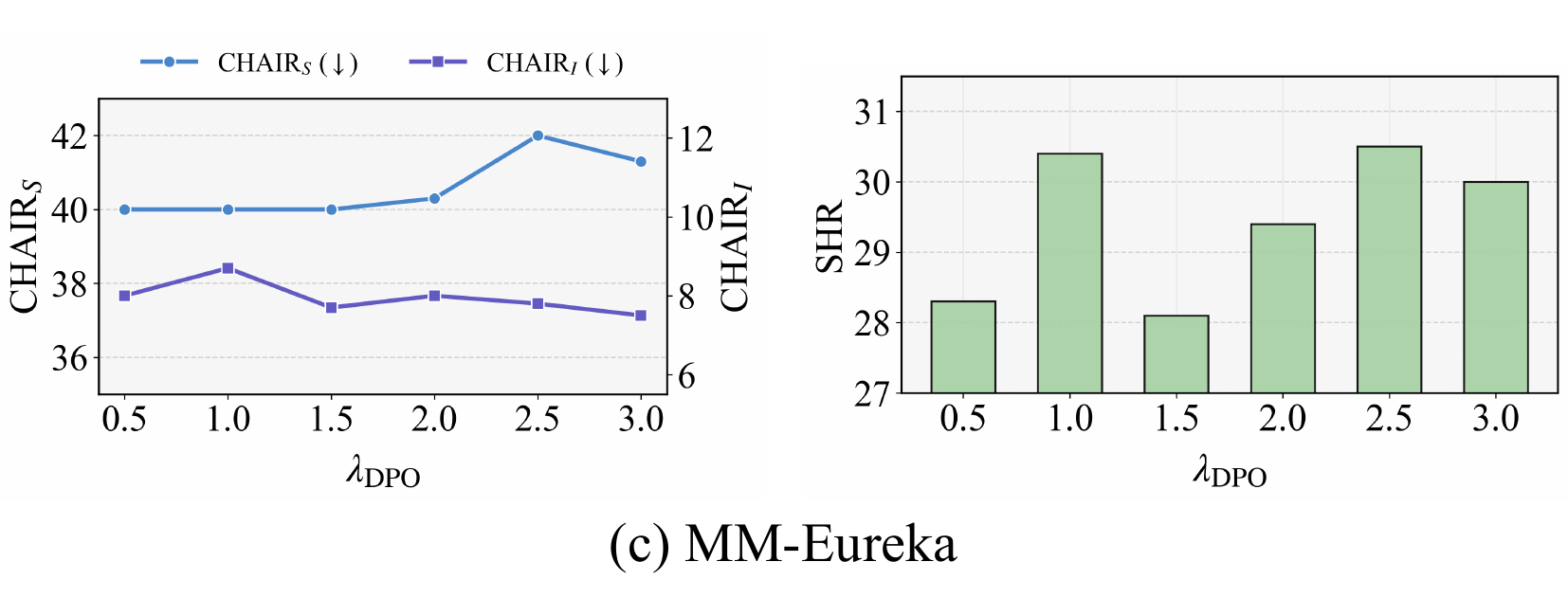}
    \vspace{0.5em}
    \includegraphics[width=\linewidth]{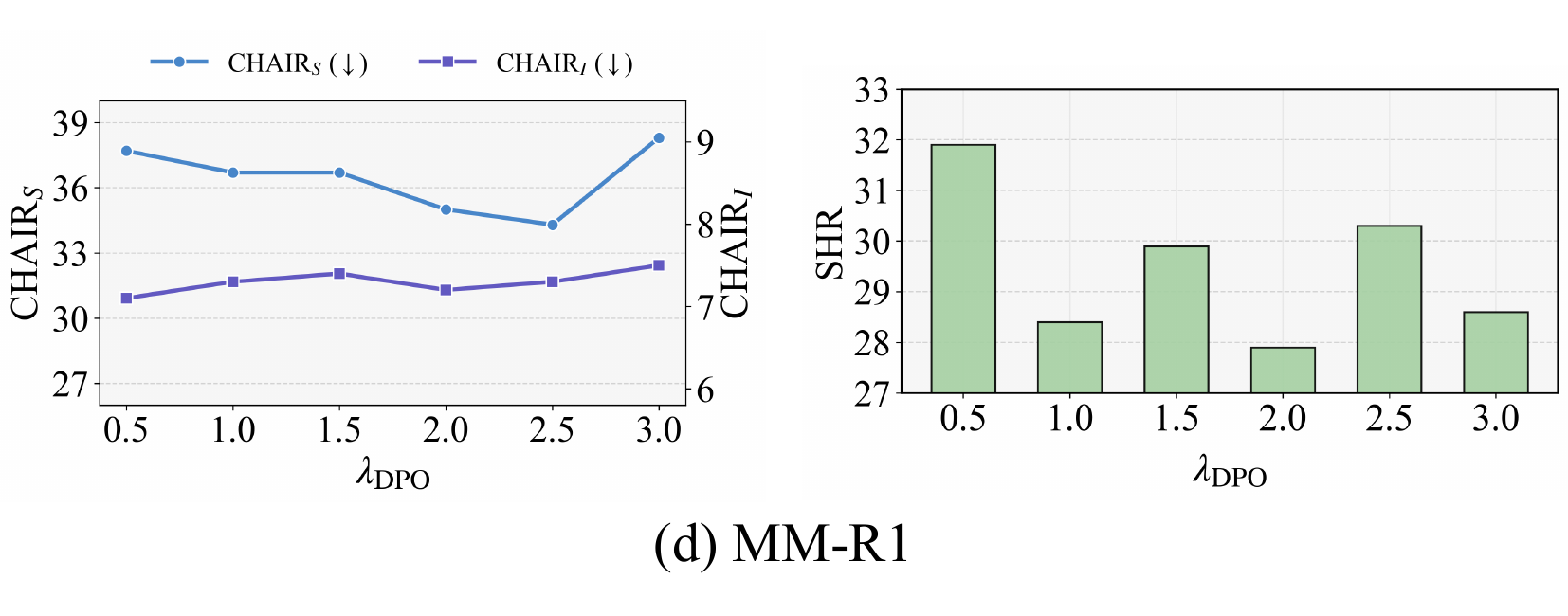}
    \vspace{0.5em}
    \includegraphics[width=\linewidth]{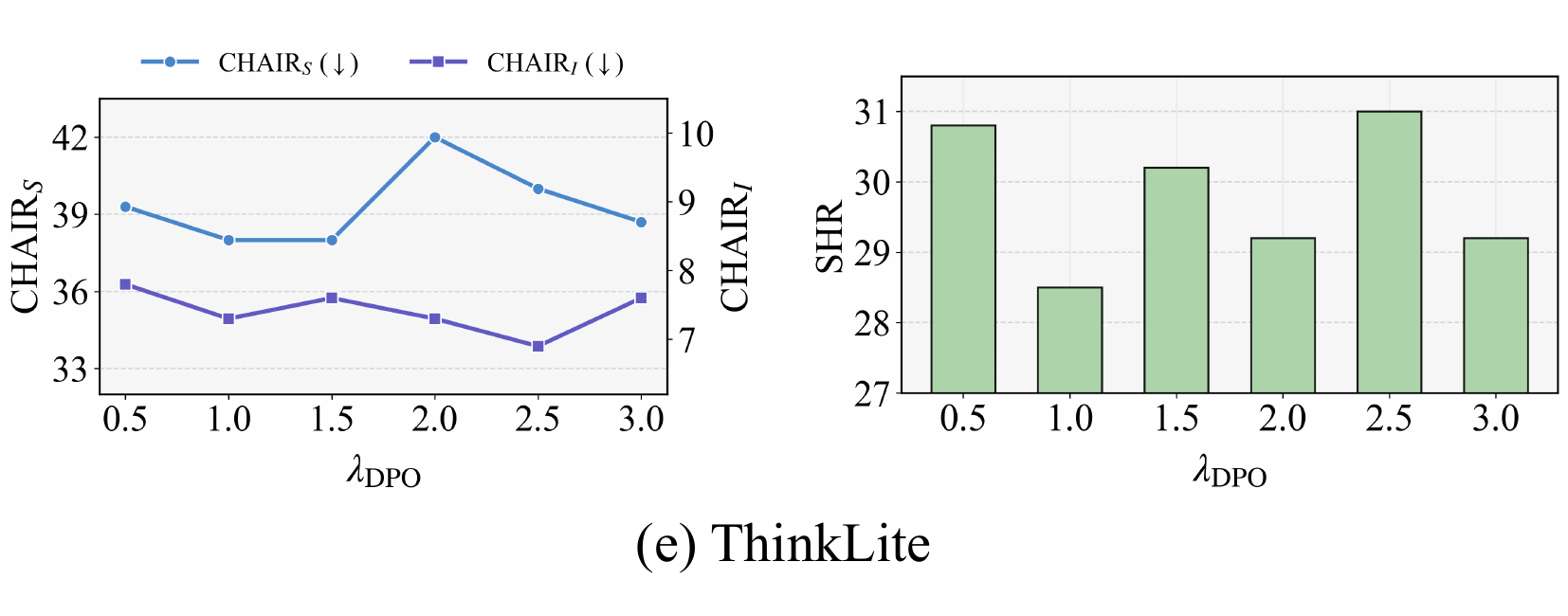}
    \caption{
    Ablation study on the weight factor $\lambda_{\mathrm{DPO}}$ of the $\mathcal{L}_{\mathrm{DPO}}$ terms.}
    \label{fig:lambda_dpo_loss_all}
    \vspace{-2.4em}
\end{wrapfigure}

\textbf{Ablation on the hyperparameter $\lambda_{\mathrm{DPO}}$}.
The value of $\lambda_{\mathrm{DPO}}$ is a critical hyperparameter as it controls the contribution of $\mathcal{L}_{\mathrm{DPO}}$ during optimization. We evaluate the performance under various values of $\lambda_{\mathrm{DPO}}$ in Figure \ref{fig:lambda_dpo_loss_all} to select the optimal value for each MRLM. The results reveal that different models exhibit distinct optimal choices of $\lambda_{\mathrm{DPO}}$. 
% For instance, Orsta-R1 achieves its lowest SHR with $\lambda_{\mathrm{DPO}}$ of 2.0. Despite this variability, the optimal values of $\lambda_{\mathrm{DPO}}$ are largely concentrated in the range of 1.5–2.0. 
In particular, we comprehensively consider the performance on both benchmarks to select the optimal values $[0.5, 2.0, 1.5, 2.0, 1.0]$ for R1-Onevision, Orsta-R1, MM-Eureka, MM-R1, ThinkLite, respectively.

\section{Related Work}
\label{sec:related_work}
\subsection{Multimodal Large Reasoning Models}
% 先说reasoning机制很重要，让LLM更强。然后过渡到多模态，介绍下MLRM，比如是如何得到的（构造数据，训练算法），然后引入代表性的MLRM（可以是我们文章中考虑的那些）。这部分可以参考之前做MLRM评估的那几篇的内容和写法。
Reasoning has become a prevalent paradigm in LLMs, as the thinking process can significantly enhance models' capabilities of complex problem solving with structured intermediate inference steps \cite{wei2022chain, kojima2022large, wang2022self}. 
Recent studies extend this paradigm to multimodal models \cite{fang2025one} and introduce Multimodal Large Reasoning Models (MLRMs) that conduct an extra reasoning step \cite{lu2022learn, zhang2023multimodal}. To acquire such capabilities, prevalent approaches typically incorporate CoT supervision via supervised fine-tuning (SFT) or reinforcement learning (RL). Early works, including RLHF-V \cite{yu2024rlhf}, LLaVA-Reasoner \cite{zhang2025improve}, and Insight-V \cite{dong2025insight}, primarily leveraged large-scale CoT-style datasets combined with preference optimization to align model behaviors with reliable reasoning patterns. Recent MLRMs typically follow two training paradigms: (i) two-stage SFT + RL strategy (e.g., R1-OneVision \cite{yang2025r1}). (ii) direct large-scale RL approach (e.g., ThinkLite-VL \cite{wang2025sota}, MM-Eureka \cite{meng2025mm}). During the RL stage, the powerful Group Relative Policy Optimization (GRPO) \cite{guo2025deepseek} is commonly adopted as an effective optimization algorithm for establishing models' reasoning capability. Notably, while models are encouraged to generate reasoning traces to tackle complex tasks, excessively long CoT chains often introduce substantial redundancy. This redundancy not only increases computational overhead but also amplifies error accumulation within the reasoning process, which instead impairs model performance \cite{chen2024not, li2025thinkless}.

\subsection{Hallucination in MLRMs}
% 介绍一下幻觉问题，说幻觉问题一直存在于多模态大模型，然后MLRM中这一问题加剧了。幻觉的表现是啥，现有的对MLRM的幻觉研究是做了些什么。强调它们并没有充分解决MLRM在通用多模态任务上的幻觉问题。
Multimodal hallucination, which indicates that generated responses are semantically coherent but contradict the visual input, has been a notorious challenge for multimodal models. 
The hallucination issue becomes even more prominent in MLRMs as longer reasoning traces shift attention away from image-grounded evidence, amplifying MLRMs' reliance on language priors during generation. \cite{liu2025more} designs a systematic benchmark to evaluate the hallucination behaviors of MLRMs and investigate several influential factors. \cite{dong2025mirage} focuses on self-contradictory hallucinations in reasoning chains and introduces a dedicated benchmark that targets reasoning-stage failures. These findings suggest that hallucinations in MLRMs are tightly coupled with the quality and structure of intermediate reasoning traces.
Existing mitigation strategies for regular MLLMs can be broadly grouped into training-free and training-based approaches. Training-free methods generally utilize prompting strategies, post-hoc correction, and improved decoding strategies (\textit{e.g.}, contrastive decoding \cite{fang2025your}) to mitigate visual inconsistencies during inference \cite{xing2024mitigating, yin2024woodpecker, huang2024opera, fang2025grounding}. Training-based approaches enhance output faithfulness by modifying the training pipeline via data curation, SFT, and reinforcement learning \cite{yu2024hallucidoctor, liu2023mitigating, jiang2024hallucination, yu2025rlaif, yang2025mitigating, kong2026reasoning, yang2026looking}. 
% and measures faithfulness along the CoT. 

Despite the progress, mitigation strategies for MLRMs in general tasks remain largely unexplored. Existing algorithms are primarily designed for conventional MLLMs and do not account for the distinctive reasoning properties of MLRMs. Consequently, how to utilize the unique reasoning mechanism to improve visual faithfulness in MLRMs remains an open problem. In this paper, we investigate the underlying hallucination causes and present the first training-based framework tailored to MLRMs for effective hallucination mitigation.

\section{Limitations and Future Work}
The CoT compression strategy in our framework is relatively simple and intuitive, which directly removes tokens deemed unimportant by a pretrained LLMlingua-2. Although this approach already yields significant improvements without introducing any additional information, a potentially more effective alternative would involve incorporating reinforcement learning (RL) to encourage the model to generate shorter yet complete CoTs or even adaptively penalizing the length of the reasoning chain specific to the input question and a predefined compression ratio.

Moreover, positive samples used in our optimization are refined by Qwen3-VL-30B-Instruct, indicating that the quality of training data heavily depends on this MLLM’s correction capability. While we observe that Qwen3-VL successfully corrects hallucinations in most cases, it still fails to provide satisfactory revisions for a certain subset of samples. Therefore, future work could focus on further improving the quality of preference data, \textit{e.g.}, by introducing a dedicated hallucination-correction module or by leveraging a multi-agent framework for collaborative refinement.

\section{Impact Statement}
% Authors are required to include a statement of the potential broader impact of their work, including its ethical aspects and future societal consequences. This statement should be in an unnumbered section at the end of the paper (co-located with Acknowledgements – the two may appear in either order, but both must be before References), and does not count toward the paper page limit. In many cases, where the ethical impacts and expected societal implications are those that are well established when advancing the field of Machine Learning, substantial discussion is not required, and a simple statement such as the following will suffice: “This paper presents work whose goal is to advance the field of Machine Learning. There are many potential societal consequences of our work, none which we feel must be specifically highlighted here.” The above statement can be used verbatim in such cases, but we encourage authors to think about whether there is content which does warrant further discussion, as this statement will be apparent if the paper is later flagged for ethics review.
This paper makes an early step toward mitigating the serious hallucination issue in MLRMs. The goal is to advance the field of large language models by improving the reliability of MLRM reasoning and reducing hallucinated outputs. There are many potential societal consequences of this line of research, most of which are consistent with those commonly associated with improving the trustworthiness of multimodal language models, and none that we believe require specific discussion here.

\section{Visualization Results}
\label{app:visual}
% 两类分析，一类是tokenskip裁剪后的，一类是幻觉相关的
% 加一句引言，说提供两类视觉分析，一个是裁剪后的，另一个是定性地揭示我们在幻觉缓解上的优越性。
% This section presents visualization results to provide a more intuitive understanding. We first show results of token pruning to demonstrate the rationality of our compression strategy. Then, we provide qualitative comparisons that confirm the superiority of C3PO in mitigating hallucinations.
\subsection{Visualization of pruned CoTs.}

\label{app:visual_pruned_cot}
\begin{figure}[t]
\begin{center}
\includegraphics[width=\linewidth]{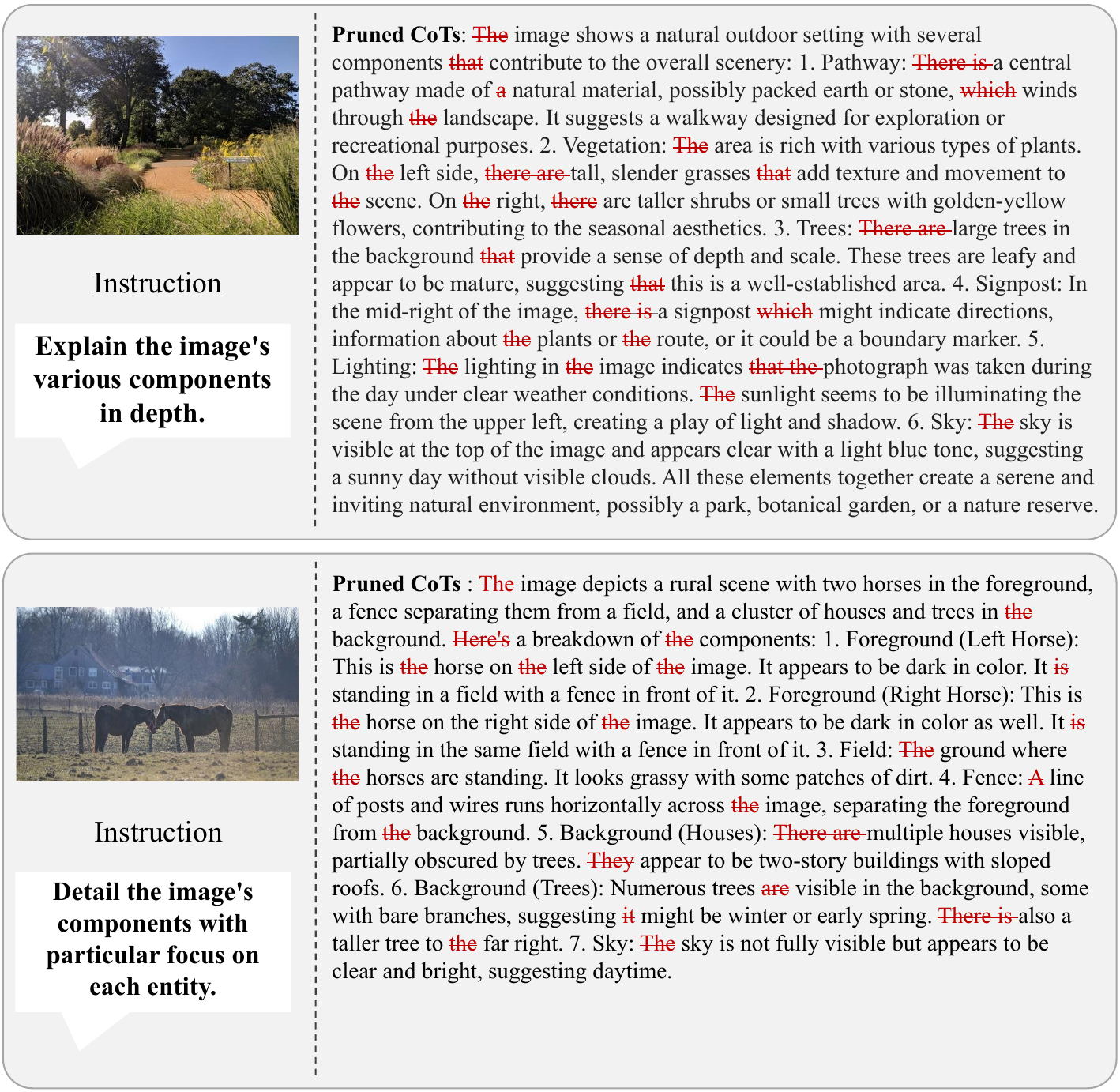}
\end{center}
\caption{Visualization of the pruned CoTs.}
\label{fig:tokenskip_vis}
\end{figure}

We visualize the pruned reasoning traces at a mask ratio of $10\%$ in Figure~\ref{fig:tokenskip_vis}. The text marked in red indicates tokens identified as low-importance and are removed during the compression phase. As observed, the removed tokens are predominantly redundant function words (\textit{e.g.}, “that”, “is”, “the”) that contribute little semantic content and are irrespective of the final response. By stripping away these low-information redundancies, our approach forms a more compact and signal-efficient CoT representation that retains task-relevant information while suppressing noise.

\subsection{Visualization of Hallucination Mitigation.}
\begin{figure}[t]
\begin{center}
\includegraphics[width=\linewidth]{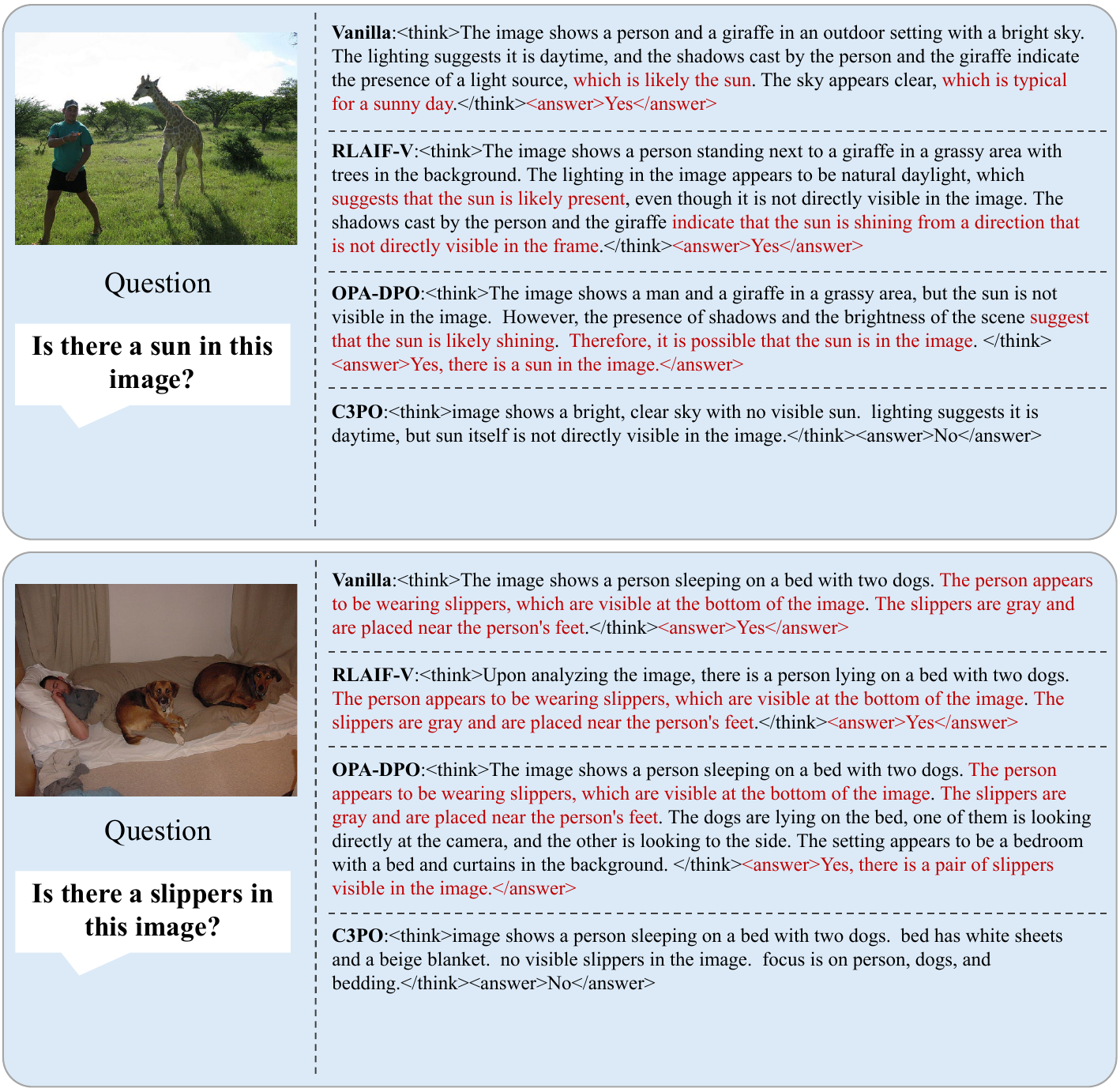}
\end{center}
\caption{Visualization results comparing our C3PO and other methods. Hallucinations are marked in \textcolor{red}{red}.}
\label{fig:hallucination_examples1}
\end{figure}

\begin{figure}[t]
\begin{center}
\includegraphics[width=\linewidth]{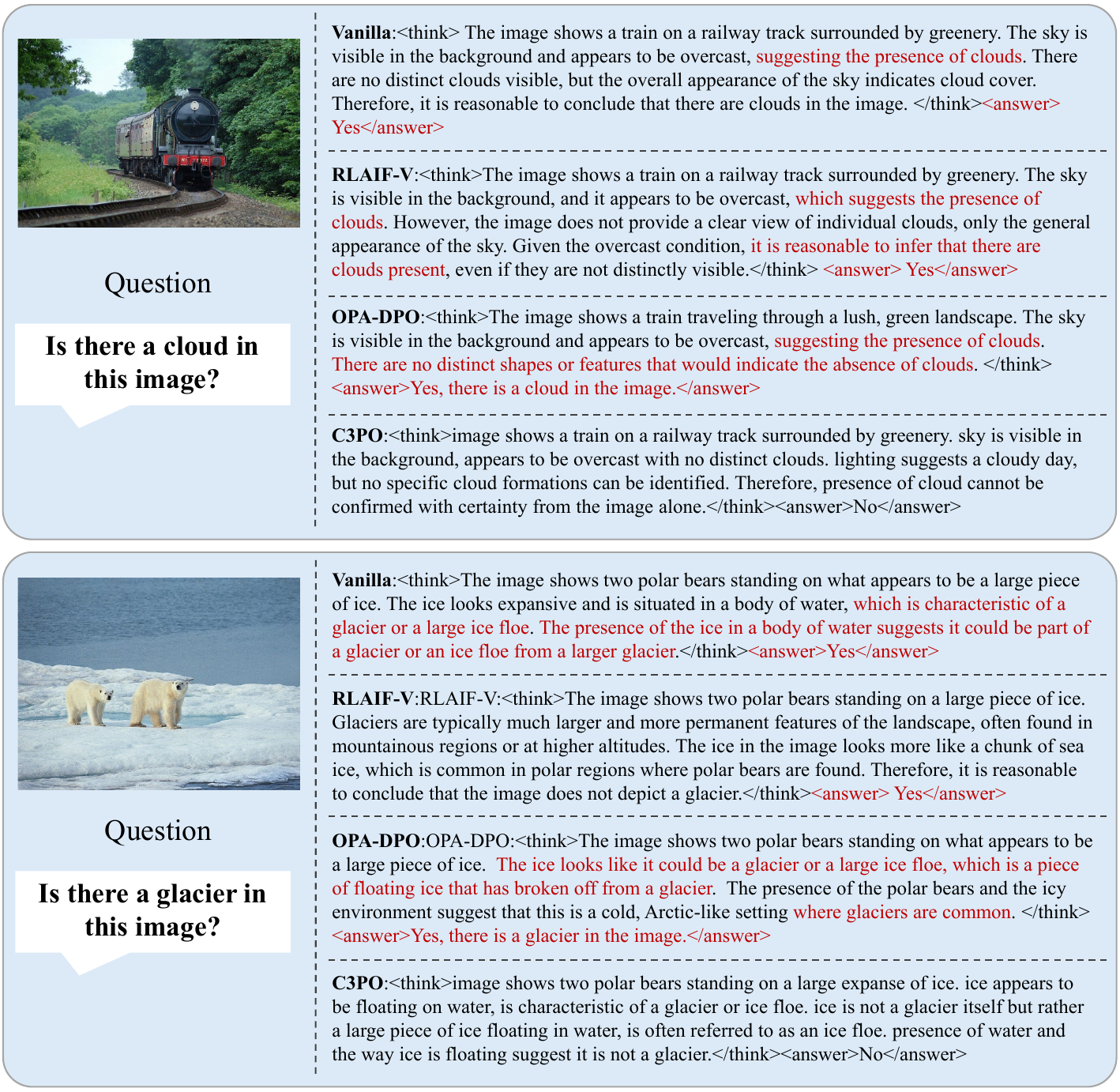}
\end{center}
\caption{Visualization results comparing our C3PO and other methods. Hallucinations are marked in \textcolor{red}{red}.}
\label{fig:hallucination_examples2}
\end{figure}

The qualitative comparisons in Figure~\ref{fig:hallucination_examples1} and Figure~\ref{fig:hallucination_examples2} reveal that baseline methods tend to exhibit hallucinations due to unreliable and low-quality reasoning. In contrast, C3PO conducts compact and reliable thinking and successfully provides the correct answer, effectively suppressing hallucinations.

\label{app:hallucination_examples}

\newpage
\clearpage 
\section*{NeurIPS Paper Checklist}

\begin{enumerate}

\item {\bf Claims}
    \item[] Question: Do the main claims made in the abstract and introduction accurately reflect the paper's contributions and scope?
    \item[] Answer: \answerYes{}  % Replace by \answerYes{}, \answerNo{}, or \answerNA{}.
    \item[] Justification: We state our contributions and research scope in the claims presented in the abstract and introduction, all of which are supported by theoretical foundation or extensive experiments.

\item {\bf Limitations}
    \item[] Question: Does the paper discuss the limitations of the work performed by the authors?
    \item[] Answer: \answerYes{} % Replace by \answerYes{}, \answerNo{}, or \answerNA{}.
    \item[] Justification: We present the limitations of our paper in section E of the Appendix.

\item {\bf Theory assumptions and proofs}
    \item[] Question: For each theoretical result, does the paper provide the full set of assumptions and a complete (and correct) proof?
    \item[] Answer: \answerYes{} % Replace by \answerYes{}, \answerNo{}, or \answerNA{}.
    \item[] Justification: We have provided full proof for the theorems in our paper.

    \item {\bf Experimental result reproducibility}
    \item[] Question: Does the paper fully disclose all the information needed to reproduce the main experimental results of the paper to the extent that it affects the main claims and/or conclusions of the paper (regardless of whether the code and data are provided or not)?
    \item[] Answer: \answerYes{} % Replace by \answerYes{}, \answerNo{}, or \answerNA{}.
    \item[] Justification: All the information of experiments are provided in Section 4 and Appendix B.

\item {\bf Open access to data and code}
    \item[] Question: Does the paper provide open access to the data and code, with sufficient instructions to faithfully reproduce the main experimental results, as described in supplemental material?
    \item[] Answer: \answerYes{} % Replace by \answerYes{}, \answerNo{}, or \answerNA{}.
    \item[] Justification: We have submitted our code in the supplementary material and will open-source the code and models.

\item {\bf Experimental setting/details}
    \item[] Question: Does the paper specify all the training and test details (e.g., data splits, hyperparameters, how they were chosen, type of optimizer) necessary to understand the results?
    \item[] Answer: \answerYes{} % Replace by \answerYes{}, \answerNo{}, or \answerNA{}.
    \item[] Justification: We provide exhaustive training and test details in our main paper and code.

\item {\bf Experiment statistical significance}
    \item[] Question: Does the paper report error bars suitably and correctly defined or other appropriate information about the statistical significance of the experiments?
    \item[] Answer: \answerYes{} % Replace by \answerYes{}, \answerNo{}, or \answerNA{}.
    \item[] Justification: We provide result analysis across varying values of hyperparameters.

\item {\bf Experiments compute resources}
    \item[] Question: For each experiment, does the paper provide sufficient information on the computer resources (type of compute workers, memory, time of execution) needed to reproduce the experiments?
    \item[] Answer: \answerYes{} % Replace by \answerYes{}, \answerNo{}, or \answerNA{}.
    \item[] Justification: We present these details in the Appendix.
    
\item {\bf Code of ethics}
    \item[] Question: Does the research conducted in the paper conform, in every respect, with the NeurIPS Code of Ethics \url{https://neurips.cc/public/EthicsGuidelines}?
    \item[] Answer:\answerYes{} % Replace by \answerYes{}, \answerNo{}, or \answerNA{}.
    \item[] Justification: The research conducted in the paper conform every respect of the NeurIPS Code of Ethics.

\item {\bf Broader impacts}
    \item[] Question: Does the paper discuss both potential positive societal impacts and negative societal impacts of the work performed?
    \item[] Answer: \answerYes{} % Replace by \answerYes{}, \answerNo{}, or \answerNA{}.
    \item[] Justification: We provide the discussion about societal impacts in Appendix F.
    
\item {\bf Safeguards}
    \item[] Question: Does the paper describe safeguards that have been put in place for responsible release of data or models that have a high risk for misuse (e.g., pre-trained language models, image generators, or scraped datasets)?
    \item[] Answer: \answerNA{} % Replace by \answerYes{}, \answerNo{}, or \answerNA{}.
    \item[] Justification: The paper poses no such risks.

\item {\bf Licenses for existing assets}
    \item[] Question: Are the creators or original owners of assets (e.g., code, data, models), used in the paper, properly credited and are the license and terms of use explicitly mentioned and properly respected?
    \item[] Answer: \answerYes{} % Replace by \answerYes{}, \answerNo{}, or \answerNA{}.
    \item[] Justification: The license and terms of use are properly respected.

\item {\bf New assets}
    \item[] Question: Are new assets introduced in the paper well documented and is the documentation provided alongside the assets?
    \item[] Answer: \answerNA{} % Replace by \answerYes{}, \answerNo{}, or \answerNA{}.
    \item[] Justification: No new assets are introduced except code.

\item {\bf Crowdsourcing and research with human subjects}
    \item[] Question: For crowdsourcing experiments and research with human subjects, does the paper include the full text of instructions given to participants and screenshots, if applicable, as well as details about compensation (if any)? 
    \item[] Answer: \answerNA{} % Replace by \answerYes{}, \answerNo{}, or \answerNA{}.
    \item[] Justification: The paper does not involve crowdsourcing nor research with human subjects.

\item {\bf Institutional review board (IRB) approvals or equivalent for research with human subjects}
    \item[] Question: Does the paper describe potential risks incurred by study participants, whether such risks were disclosed to the subjects, and whether Institutional Review Board (IRB) approvals (or an equivalent approval/review based on the requirements of your country or institution) were obtained?
    \item[] Answer: \answerNA{} % Replace by \answerYes{}, \answerNo{}, or \answerNA{}.
    \item[] Justification: This paper does not involve crowdsourcing nor research with human subjects.

\item {\bf Declaration of LLM usage}
    \item[] Question: Does the paper describe the usage of LLMs if it is an important, original, or non-standard component of the core methods in this research? Note that if the LLM is used only for writing, editing, or formatting purposes and does \emph{not} impact the core methodology, scientific rigor, or originality of the research, declaration is not required.
    %this research? 
    \item[] Answer: \answerNA{} % Replace by \answerYes{}, \answerNo{}, or \answerNA{}.
    \item[] Justification: In this paper, we does not utilize LLMs for core method development nor as any important, original, or non-standard components.

\end{enumerate}

\end{document}